\documentclass{article} 
\usepackage{iclr2021_conference,times}

\usepackage{amsmath,amsfonts,bm}

\def\eqref#1{equation~\ref{#1}}

\def\1{\bm{1}}

\DeclareMathAlphabet{\mathsfit}{\encodingdefault}{\sfdefault}{m}{sl}
\SetMathAlphabet{\mathsfit}{bold}{\encodingdefault}{\sfdefault}{bx}{n}

\usepackage{hyperref}
\usepackage{url}

\usepackage{todonotes}

\usepackage{algorithm}
\usepackage{algorithmic}

\usepackage{booktabs} 
\usepackage{tabularx}
\usepackage{tabulary}
\usepackage{multirow}
\newcolumntype{Y}{>{\centering\arraybackslash}X}

\usepackage{subcaption}

\usepackage{amssymb}

\usepackage{bm}
\usepackage{chngcntr}
\usepackage{float}
\usepackage{stfloats} 

\usepackage{xspace}
\newcommand{\fmix}{FMix\xspace}

\newcommand{\iocc}{iOcclusion\xspace}
\newcommand{\cutocc}{CutOcclusion\xspace}
\newcommand{\mixup}{MixUp\xspace}
\newcommand{\cutmix}{CutMix\xspace}
\newcommand{\cutout}{CutOut\xspace}

\newcommand{\cifar}[1]{CIFAR-{#1}\xspace}

\newcommand{\imagenet}{ImageNet\xspace}

\usepackage[export]{adjustbox}

\title{On Pitfalls of Measuring Occlusion \\Robustness through Data Distortion}

\author{Antonia Marcu \\
Vision, Learning and Control Group\\
University of Southampton\\
\texttt{am1g15@soton.ac.uk} \\
}

\iclrfinalcopy 
\begin{document}

\maketitle

\begin{abstract}
    
    Over the past years, the crucial role of data has largely been shadowed by the field's focus on architectures and training procedures.
    We often cause changes to the data without being aware of their wider implications. 
    In this paper we show that distorting images without accounting for the artefacts introduced leads to biased results when establishing occlusion robustness.
    To ensure models behave as expected in real-world scenarios, we need to rule out the impact added artefacts have on evaluation.
    We propose a new approach, \iocc, as a fairer alternative for applications where the possible occluders are unknown.
    
\end{abstract}

\section{Introduction}
\label{sec:intro}

Correctly assessing the ability of a model to perform despite input alterations is crucial for obtaining reliable systems.
Despite their incontestable success in a number of visual tasks, deep models are not fully trusted for real-world applications because of their sensitivity to input changes.
This is an active area of research and proposed solutions are both at a data \citep{devries2017improved, yun2019cutmix} as well as algorithm \citep{globerson2006nightmare, kortylewski2020compositional, zhu2019robustness, xu2020robust} level. 
A widely adopted method for measuring occlusion robustness is through the accuracy obtained after superimposing a rectangular patch on an image \citep{chun2020empirical, yun2019cutmix, fawzi2016measuring, yun2019cutmix, zhong2020random, kokhlikyan2020captum}. 
In this paper, we refer to this approach as \cutocc due to its similarity to the \cutout augmentation \citep{devries2017improved} 
and argue it can be misleading, especially in comparative studies and applications where there is no prior knowledge about the exact shape of the possible occluders. 
Subsequently, to address the unfairness observed, we introduce \iocc, a measure that could form the baseline for future robustness studies.

\section{Is occlusion robustness measured fairly?} \label{sec:occlusion}

To verify the existence of the aforementioned bias, we need to compare models with different behaviours. To do this in a  controlled manner, we make use of data augmentation.
We focus on two popular mixed-sample augmentations, \mixup~\citep{zhang2017mixup} and \cutmix~\citep{yun2019cutmix}.
\mixup linearly interpolates between two images to obtain a new training example, while \cutmix superimposes onto a sample a rectangular region taken from another image. We also use \fmix due to its irregularly shaped masks sampled from Fourier space, which will play an important role in our analysis.
See Section \ref{sup:visual_examples} of the Supplementary Material for samples obtained with each method.
We train PreAct-ResNet18~\citep{he2016identity} and VGG~\citep{simonyan2014very} models on \cifar{10/100}~\citep{krizhevsky2009learning}, FashionMNIST~\citep{xiao2017fashion} with these above augmentations. See Section~\ref{sup:exp_details} for experimental details. 

For the four types of models (basic, \mixup, \fmix and \cutmix) we look at the increase in misclassifications per category, when presented with CutOccluded images.
That is, from the number of incorrect predictions of a model evaluated on modified data, we subtract the incorrect predictions when testing on original data.
If one of the classes has a significant increase, this indicates that the distortion introduces features the model associates with that class.
For all data sets, the basic model tends to wrongly predict a specific class, while between the augmented models at least one is invariant to this distortion.
For example, on \cifar{10}, the basic and \mixup models tend to misclassify CutOccluded images as ``Truck'' (Figure~\ref{fig:truck}).
This is not at all surprising, given that the strong horizontal and vertical edges are highly indicative of this class.
Section~\ref{sup:wrongPred} contains more examples.

Admittedly, this experiment is not meant to provide a rigorous account of the bias and a comparison between the precise extent of the bias identified in various models is not possible at this stage.
Nonetheless, it is sufficient to capture the broad phenomenon.
Thus, by occluding images using a particularly shaped patch one implicitly measures a model's affinity to certain features, albeit those features might be discriminative.
This deems such a method inappropriate for fairly assessing robustness. Moreover, we identify similar issues in the case of texture bias identification and augmentation analyses. In the interest of space, we omit the results from the current paper, but believe this observation calls for more principled uses of data distortion in model evaluation.

A related observation was made independently of us by \citet{hooker2019benchmark} who note the pitfalls of image manipulation in the context of interpretability methods.
They focus on the distributional shift induced when removing image regions to determine feature importance.
That is, they point out that when simply superimposing uniform patches over image features, it is difficult to asses how much of the reduction in accuracy is caused by the absence of those features and how much is due to images becoming out of distribution.
To address this, the most important features both on train and test data are masked out, closing the gap between the two sets. 
They then train and respectively evaluate models on the newly generated images.
Here we study a similar problem. 
We are interested in ruling out the overlap between the model's learned representations and the information that is introduced by the robustness measurement method.
Unlike for interpretability methods, the subject of occlusion studies is the model itself and, as such, training with a modified version of the data is not a viable option.
In the following section we explore ways of overcoming this bias when measuring occlusion robustness.


\subsection{\iocc}

\begin{figure}
\begin{minipage}{0.48\linewidth}
    \centering
    \includegraphics[width=\linewidth]{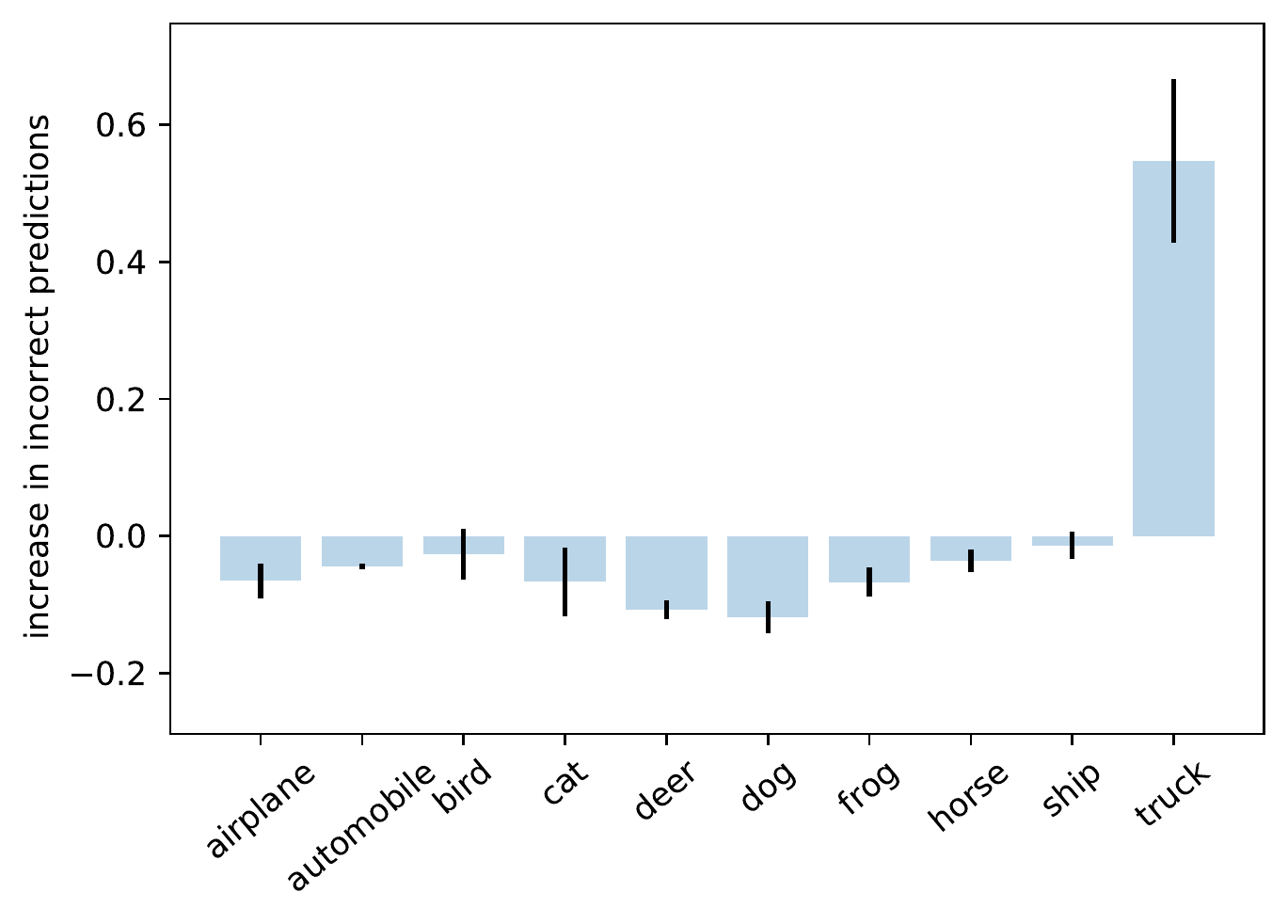}
    \caption{Difference in incorrect predictions for the basic model on CutOccluded images.}
    \label{fig:truck}
\end{minipage}
\hfill
\begin{minipage}{0.48\linewidth}
    \centering
    \includegraphics[width=0.9\linewidth]{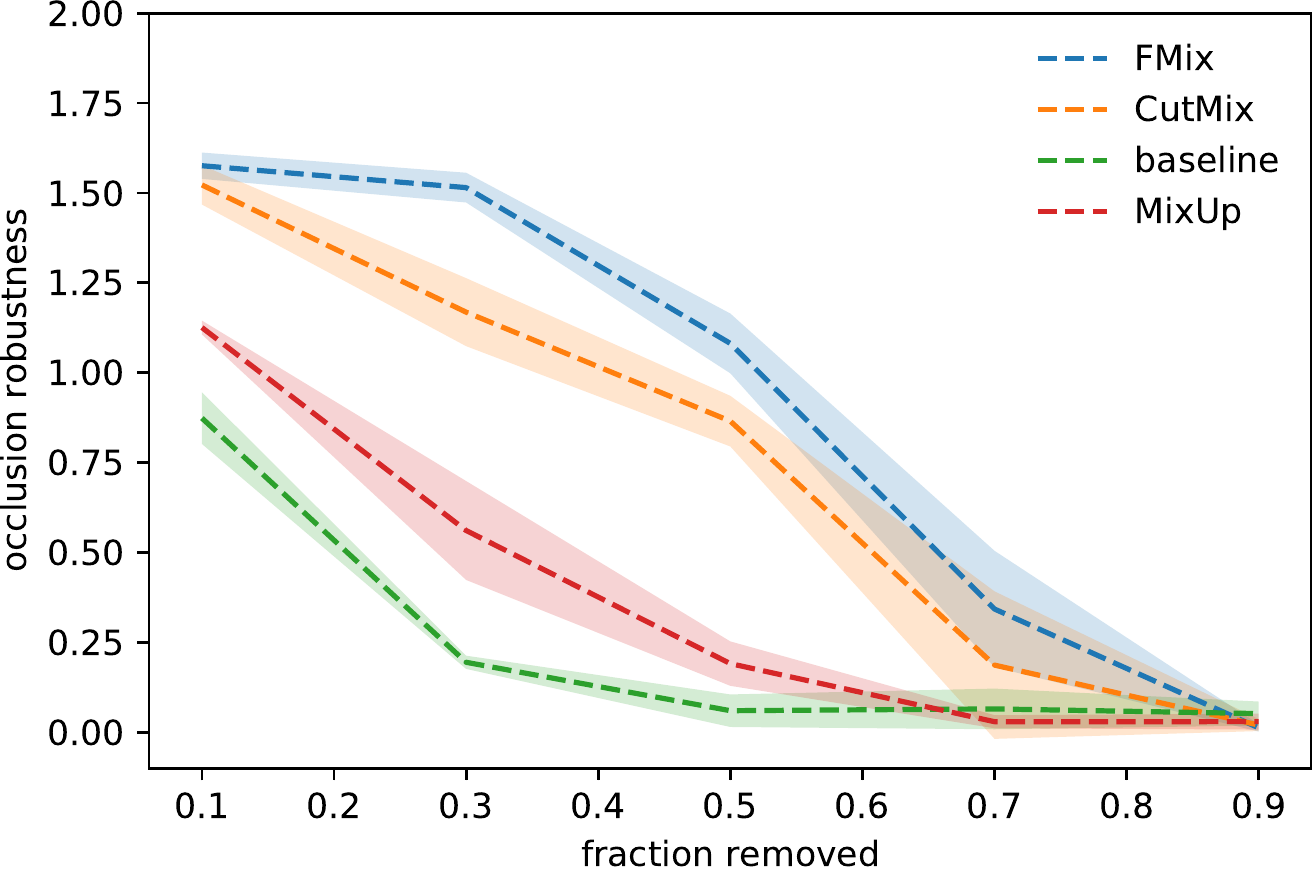}
    \caption{Occlusion robustness for varying fractions of image obstruction as measured with \iocc for the four types of models trained on \cifar{10}.}
    \label{fig:iocc_nomix}
\end{minipage}

\end{figure}

We propose a simple measure that aims to decouple the machine's edge bias from the occlusion robustness, which we refer to as ``invariant Occlusion'' (\iocc).
Invariant Occlusion reflects the change in the interplay between performance on seen and unseen data. Formally,
\begin{equation}
    \iocc_{i} = \left| \frac{ \mathcal{A}{(\mathcal{D}^{i}_{train})}- \mathcal{A}{(\mathcal{D}^{i}_{test})}}{\mathcal{A}{(\mathcal{D}_{train})}- \mathcal{A}{(\mathcal{D}_{test})}} \right|,    
    \label{eq:iocc}
\end{equation}
where $\mathcal{A}(\mathcal{D})$ denotes the accuracy on a given data set $\mathcal{D}$, and $\mathcal{D}^{i}$ is the data set resulting from removing $i\%$ pixels of each image.
The intuition is that on train data robust models are less sensitive to the artefacts of the occlusion policy for small levels of occlusion, resulting in a large difference in accuracy from that on unseen data.
The performance of both train and test gets close to random as the percentage of sample occlusion approaches 90\% and we expect the gap to fall off quicker for less robust models. 
This change in interplay is taken with respect to the generalisation gap of the model. Thus, the denominator plays a normalisation role such that the quality of the model fit in itself does not interfere with the robustness measure. 
In this paper we choose to generate masks using Grad-CAM \citep{selvaraju2017grad}, such that the area with most salient i\% pixels is covered.
It must be noted that this method implicitly assumes there could be multiple occluders.
We also experiment with using rectangular or Fourier-sampled masks and conclude that although random masking makes the process noisier, the exact choice of masking method is of secondary importance.

In Figure~\ref{fig:iocc_nomix}, we evaluate the robusteness using our method.
Note that unlike the typical measure, \iocc is not necessarily a proper fraction.
The higher robustness of \fmix over \cutmix is justified through the sparsity of \fmix masking, as well as the way the two determine the size of the occluding patch. 
Although they sample the size from the same distribution, in \cutmix part of the rectangle can be outside the image, which results in less occluded images overall. 
Thus, we will generally use the basic and \fmix models as reference for least and most robust respectively.

\subsection{Evaluation}

\begin{table*}[t]
    \centering
    \caption{\iocc and \cutocc at a 30\% obstruction level. Note that there is a difference in scale and the two should not be directly compared. We are interested in how the methods situate the models with respect to each other. When measuring the robustness with \cutocc, RM3 appears significantly less robust than \cutmix due to its sensitivity to patching with rectangles, while \iocc highlights the robustness specific to \fmix-like masks.}\label{tab:3msks}
    \begin{tabulary}{\linewidth}{lLLLLLL}
    \toprule
                & basic & \mixup & \cutmix & \fmix & RM & RM3 \\
    \cutocc  & $40.59_{\pm2.27}$ & $52.64_{\pm1.77}$ & $82.43_{\pm 1.60}$ & $82.67_{\pm 0.29}$ & $53.50_{4.76}$ & $66.43_{\pm 6.33}$ \\ 
    \iocc & $0.19_{\pm0.01}$ & $0.56_{\pm0.13}$ & $1.16_{\pm 0.09}$ & $1.51_{\pm 0.04}$ & $0.63_{\pm 0.29}$ & $1.45_{\pm 0.19}$\\  
    \bottomrule
    \end{tabulary}
\end{table*}

Assessing the correctness of such a measure is difficult in the absence of a basic. 
For the rest of this section we will build varied experiments to attest the validity of our method.
We being by empirically confirming \iocc rules out the edge bias in \cutocc. We would like to train with an augmentation that has irregularly shaped masks, but at the same time doesn't have the variation of \fmix, which causes it to be insensitive to \cutmix's edges.  
To this end, we create two variations of \fmix with which we train: RM (Random Mask) --- where a single mask is sampled for the entire training; and RM3 --- where for each batch one of three fixed masks is chosen uniformly at random. 
As for all our experiments, we do 5 repeats, for each one sampling different fixed masks.
As desired, unlike \fmix and \cutmix, the models trained with the fixed-mask augmentation versions are highly sensitive to the edge artefacts of \cutocc (see Section~\ref{sup:wrongPred}). 
Note that  to allow a fairer comparison to other methods, for \cutocc we do not allow the obstructing patch to lie outside the image such that the fraction removed is exact.
The results in Table~\ref{tab:3msks} are obtained for a fraction of 0.3 pixels removed. 
Our measure reflects the robustness of training with three fixed random masks (RM3), situating it closer to masking methods rather than interpolative. Furthermore, \iocc captures the large variance in training with a fixed random mask, which shows they can range from providing almost masking-MSDA robustness to worse than \mixup.

Additionally, since occlusion in real-life scenarios could also be caused by objects that have a texture of their own, an appropriate measure must not be sensitive to the occluder's pattern.
We verify this by superimposing patches from images belonging to a different data set.
For this, we perform the same operation on input as mixed-sample masking MSDAs, where the mask is given by Grad-CAM.
We choose to mask with images from different data sets to avoid the side-effects of samples belonging to two classes simultaneously.
The randomness introduced by the texture is naturally making the the process noisier.
Nonetheless, we find again \iocc to better rule out the specifics of the occluding patch, whereas \cutocc provides significantly different results to its uniform version, pushing everything together (see Section~\ref{sup:mix_nomix} of the Supplementary Material). 



\begin{figure*}
    \begin{subfigure}[t]{0.49\linewidth}
        \centering
        \includegraphics[width=\linewidth]{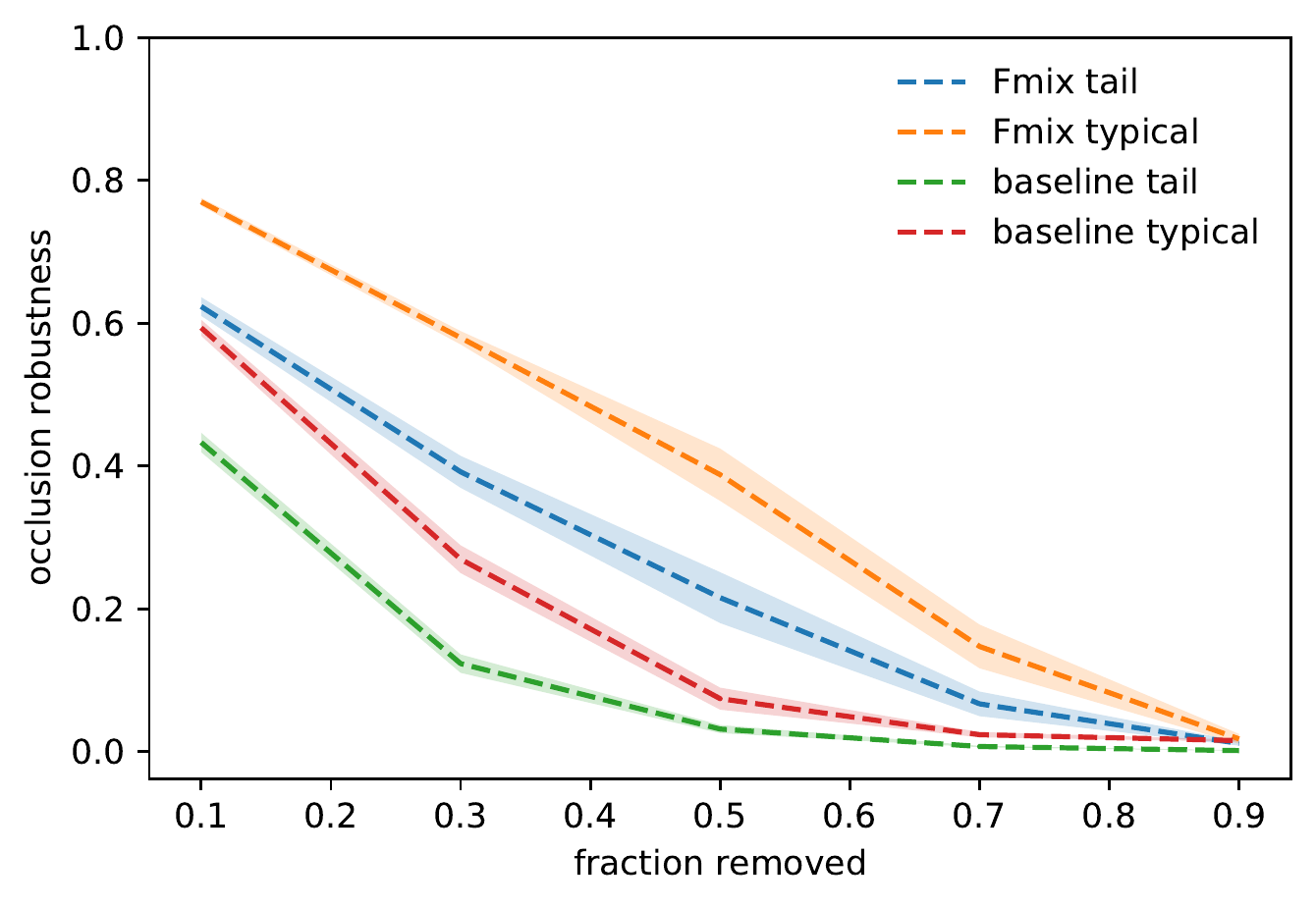}
    \end{subfigure}
    \hfill 
    \begin{subfigure}[t]{0.49\linewidth}
        \centering
        \includegraphics[width=\linewidth]{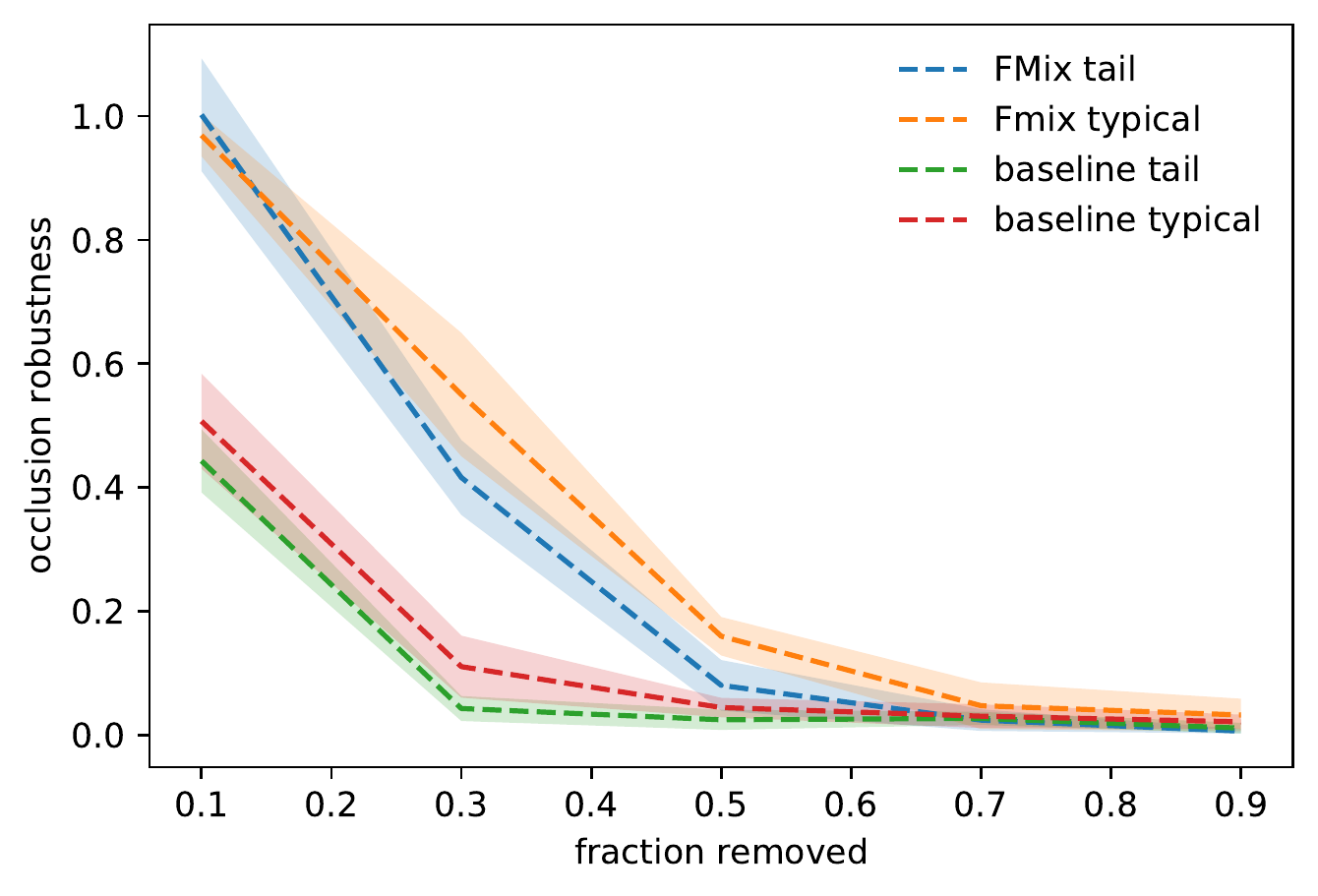}
    \end{subfigure}
    \caption{ \cutocc (left) and \iocc (right) for the basic and \fmix models on two subsets of the same data set, which we refer to as tail and typical. Evaluating the models with \iocc on the two types of samples leads robustness levels that do not differ outside the margin of error. However, \cutocc finds the models to be less robust on tail data. }
    \label{fig:tail-typ}
\end{figure*}

Another problem that occurs when purely looking at post-masking accuracy is weaker models would erroneously appear less robust. 
We show this by reversing the problem: we evaluate the same model on two subsets of the \cifar{100} data set: typical and atypical images as categorised by \citet{feldman2020neural}. 
Each tail example from the train set has a corresponding one in the test set.
\cutocc would indicate that models are significantly more robust to occluding typical examples (see Figure~\ref{fig:tail-typ}). 
However, a closer analysis makes us doubt this conclusion.
The raw accuracy on both train and test data for tail examples is lower than for the typical ones and the decrease in performance when masking out image regions is the same for the two subsets.
By definition, \iocc allows a fair comparison of robustness regardless of the overall performance of a model.

\begin{table}[t]
    \centering
    \caption{Robustness to occluding with patches covering $50\%$ of each image (for a full range of percentages, see Section~\ref{sup:rand_lbls}). The models are trained with and without masking augmentation on data with randomised labels. 
    \cutocc makes no difference between regular and augmented training.} \label{tab:rand_oriz}
    \begin{tabulary}{\linewidth}{lLLL}
    \toprule
     & basic random & \fmix random & \fmix clean \\
     \midrule
    \cutocc &$10.24_{\pm0.27}$ &$9.78_{\pm0.18}$ &$63.63_{\pm4.54}$ \\
    \iocc   &$14.63_{\pm1.12}$ & $47.94_{\pm19.84}$ & $82.36_{\pm10.06}$\\
    \bottomrule
    \end{tabulary}
\end{table}

Lastly, we train models on \cifar{10} with randomly assigned labels as is done in \citet{zhang2016understanding} with and without \fmix, until examples are memorised. 
Since all labels are corrupted, the accuracy on the test set before and after occlusion is no greater than random. 
However, the robustness of the augmentation-trained model can be seen on the training data, as captured by our metric.
On the other hand, \cutocc makes no distinction between learning with regular and augmented data (Table~\ref{tab:rand_oriz}).
Despite being such a peculiar case, it shows the comprehensiveness gained by accounting for the degradation on test data in relation to that on train.

Thus, as we evidenced through controlled experiments, there are many cases that \cutocc does not properly address.
From a model analysis perspective, correctly assessing the occlusion robustness could lead to better understanding and development of models and training procedure. 
Equally important, it has applicability for real-world deployments where no prior knowledge exists about the possible shapes of the obstructions.
The strength of the bias will depend on the data in question. 
Thus, some applications will be more heavily affected than others. 
However, we have seen that for natural images as is the case for \cifar{10} and \cifar{100} or \imagenet~\citep{russakovsky2015imagenet}, this bias does exist. 
We believe the edge artefacts are very likely to interfere with learned representations since they are such fundamental features.
Thus, accounting for the bias is necessary to ensure a correct robustness assessment.

\section{Conclusions}

Distorting data without investigating its broader effects is particularly problematic when applied in analyses, as is the case of occlusion robustness measurement.
We show the typical approach, \cutocc, is biased.
This deems it inappropriate for evaluating models for real-world applications, where there is a variety of possible occluding objects.
We propose \iocc as a basic for future robustness studies and design a number of experiments to validate it.
In a broader sense, the purpose of our paper is to encourage better practice when dealing with data distortions.

\bibliography{iclr2021_conference}

\begin{thebibliography}{21}
\providecommand{\natexlab}[1]{#1}
\providecommand{\url}[1]{\texttt{#1}}
\expandafter\ifx\csname urlstyle\endcsname\relax
  \providecommand{\doi}[1]{doi: #1}\else
  \providecommand{\doi}{doi: \begingroup \urlstyle{rm}\Url}\fi

\bibitem[Chun et~al.(2020)Chun, Oh, Yun, Han, Choe, and Yoo]{chun2020empirical}
Sanghyuk Chun, Seong~Joon Oh, Sangdoo Yun, Dongyoon Han, Junsuk Choe, and
  Youngjoon Yoo.
\newblock An empirical evaluation on robustness and uncertainty of
  regularization methods.
\newblock \emph{arXiv preprint arXiv:2003.03879}, 2020.

\bibitem[DeVries \& Taylor(2017)DeVries and Taylor]{devries2017improved}
Terrance DeVries and Graham~W Taylor.
\newblock Improved regularization of convolutional neural networks with cutout.
\newblock \emph{arXiv preprint arXiv:1708.04552}, 2017.

\bibitem[Fawzi \& Frossard(2016)Fawzi and Frossard]{fawzi2016measuring}
Alhussein Fawzi and Pascal Frossard.
\newblock Measuring the effect of nuisance variables on classifiers.
\newblock In \emph{British Machine Vision Conference (BMVC)}, number CONF,
  2016.

\bibitem[Feldman \& Zhang(2020)Feldman and Zhang]{feldman2020neural}
Vitaly Feldman and Chiyuan Zhang.
\newblock What neural networks memorize and why: Discovering the long tail via
  influence estimation.
\newblock \emph{Advances in Neural Information Processing Systems}, 33, 2020.

\bibitem[Globerson \& Roweis(2006)Globerson and Roweis]{globerson2006nightmare}
Amir Globerson and Sam Roweis.
\newblock Nightmare at test time: robust learning by feature deletion.
\newblock In \emph{Proceedings of the 23rd international conference on Machine
  learning}, pp.\  353--360, 2006.

\bibitem[Harris et~al.(2020)Harris, Marcu, Painter, Niranjan,
  Pr{\"u}gel-Bennett, and Hare]{harris2020understanding}
Ethan Harris, Antonia Marcu, Matthew Painter, Mahesan Niranjan, Adam
  Pr{\"u}gel-Bennett, and Jonathon Hare.
\newblock Understanding and enhancing mixed sample data augmentation.
\newblock \emph{arXiv preprint arXiv:2002.12047}, 2020.

\bibitem[He et~al.(2016)He, Zhang, Ren, and Sun]{he2016identity}
Kaiming He, Xiangyu Zhang, Shaoqing Ren, and Jian Sun.
\newblock Identity mappings in deep residual networks.
\newblock In \emph{European conference on computer vision}, pp.\  630--645.
  Springer, 2016.

\bibitem[Hooker et~al.(2019)Hooker, Erhan, Kindermans, and
  Kim]{hooker2019benchmark}
Sara Hooker, Dumitru Erhan, Pieter-Jan Kindermans, and Been Kim.
\newblock A benchmark for interpretability methods in deep neural networks.
\newblock In \emph{Advances in Neural Information Processing Systems}, pp.\
  9737--9748, 2019.

\bibitem[Kokhlikyan et~al.(2020)Kokhlikyan, Miglani, Martin, Wang, Alsallakh,
  Reynolds, Melnikov, Kliushkina, Araya, Yan, and
  Reblitz-Richardson]{kokhlikyan2020captum}
Narine Kokhlikyan, Vivek Miglani, Miguel Martin, Edward Wang, Bilal Alsallakh,
  Jonathan Reynolds, Alexander Melnikov, Natalia Kliushkina, Carlos Araya, Siqi
  Yan, and Orion Reblitz-Richardson.
\newblock Captum: A unified and generic model interpretability library for
  pytorch, 2020.

\bibitem[Kortylewski et~al.(2020)Kortylewski, Liu, Wang, Sun, and
  Yuille]{kortylewski2020compositional}
Adam Kortylewski, Qing Liu, Angtian Wang, Yihong Sun, and Alan Yuille.
\newblock Compositional convolutional neural networks: A robust and
  interpretable model for object recognition under occlusion.
\newblock \emph{International Journal of Computer Vision}, pp.\  1--25, 2020.

\bibitem[Krizhevsky et~al.(2009)]{krizhevsky2009learning}
Alex Krizhevsky et~al.
\newblock Learning multiple layers of features from tiny images.
\newblock 2009.

\bibitem[Russakovsky et~al.(2015)Russakovsky, Deng, Su, Krause, Satheesh, Ma,
  Huang, Karpathy, Khosla, Bernstein, et~al.]{russakovsky2015imagenet}
Olga Russakovsky, Jia Deng, Hao Su, Jonathan Krause, Sanjeev Satheesh, Sean Ma,
  Zhiheng Huang, Andrej Karpathy, Aditya Khosla, Michael Bernstein, et~al.
\newblock Imagenet large scale visual recognition challenge.
\newblock \emph{International journal of computer vision}, 115\penalty0
  (3):\penalty0 211--252, 2015.

\bibitem[Selvaraju et~al.(2017)Selvaraju, Cogswell, Das, Vedantam, Parikh, and
  Batra]{selvaraju2017grad}
Ramprasaath~R Selvaraju, Michael Cogswell, Abhishek Das, Ramakrishna Vedantam,
  Devi Parikh, and Dhruv Batra.
\newblock Grad-cam: Visual explanations from deep networks via gradient-based
  localization.
\newblock In \emph{Proceedings of the IEEE international conference on computer
  vision}, pp.\  618--626, 2017.

\bibitem[Simonyan \& Zisserman(2014)Simonyan and Zisserman]{simonyan2014very}
Karen Simonyan and Andrew Zisserman.
\newblock Very deep convolutional networks for large-scale image recognition.
\newblock \emph{arXiv preprint arXiv:1409.1556}, 2014.

\bibitem[Xiao et~al.(2017)Xiao, Rasul, and Vollgraf]{xiao2017fashion}
Han Xiao, Kashif Rasul, and Roland Vollgraf.
\newblock Fashion-mnist: a novel image dataset for benchmarking machine
  learning algorithms.
\newblock \emph{arXiv preprint arXiv:1708.07747}, 2017.

\bibitem[Xu et~al.(2020)Xu, Liu, Yang, and Niethammer]{xu2020robust}
Zhenlin Xu, Deyi Liu, Junlin Yang, and Marc Niethammer.
\newblock Robust and generalizable visual representation learning via random
  convolutions.
\newblock \emph{arXiv preprint arXiv:2007.13003}, 2020.

\bibitem[Yun et~al.(2019)Yun, Han, Oh, Chun, Choe, and Yoo]{yun2019cutmix}
Sangdoo Yun, Dongyoon Han, Seong~Joon Oh, Sanghyuk Chun, Junsuk Choe, and
  Youngjoon Yoo.
\newblock Cutmix: Regularization strategy to train strong classifiers with
  localizable features.
\newblock In \emph{Proceedings of the IEEE International Conference on Computer
  Vision}, pp.\  6023--6032, 2019.

\bibitem[Zhang et~al.(2017{\natexlab{a}})Zhang, Bengio, Hardt, Recht, and
  Vinyals]{zhang2016understanding}
Chiyuan Zhang, Samy Bengio, Moritz Hardt, Benjamin Recht, and Oriol Vinyals.
\newblock Understanding deep learning requires rethinking generalization.
\newblock In \emph{International Conference on Learning Representations},
  2017{\natexlab{a}}.

\bibitem[Zhang et~al.(2017{\natexlab{b}})Zhang, Cisse, Dauphin, and
  Lopez-Paz]{zhang2017mixup}
Hongyi Zhang, Moustapha Cisse, Yann~N Dauphin, and David Lopez-Paz.
\newblock mixup: Beyond empirical risk minimization.
\newblock \emph{arXiv preprint arXiv:1710.09412}, 2017{\natexlab{b}}.

\bibitem[Zhong et~al.(2020)Zhong, Zheng, Kang, Li, and Yang]{zhong2020random}
Zhun Zhong, Liang Zheng, Guoliang Kang, Shaozi Li, and Yi~Yang.
\newblock Random erasing data augmentation.
\newblock In \emph{AAAI}, pp.\  13001--13008, 2020.

\bibitem[Zhu et~al.(2019)Zhu, Tang, Park, Park, and Yuille]{zhu2019robustness}
Hongru Zhu, Peng Tang, Jeongho Park, Soojin Park, and Alan Yuille.
\newblock Robustness of object recognition under extreme occlusion in humans
  and computational models.
\newblock \emph{arXiv preprint arXiv:1905.04598}, 2019.

\end{thebibliography}
\bibliographystyle{iclr2021_conference}

\newpage
\appendix
\begin{center}
\Large \bf Supplementary Material
\end{center}

\section{Augmentation examples} \label{sup:visual_examples}

\begin{figure}
\begin{tabularx}{0.6\linewidth}{Xl}

    Image 1 & \multirow{5}{*}{\includegraphics[width=0.4\linewidth]{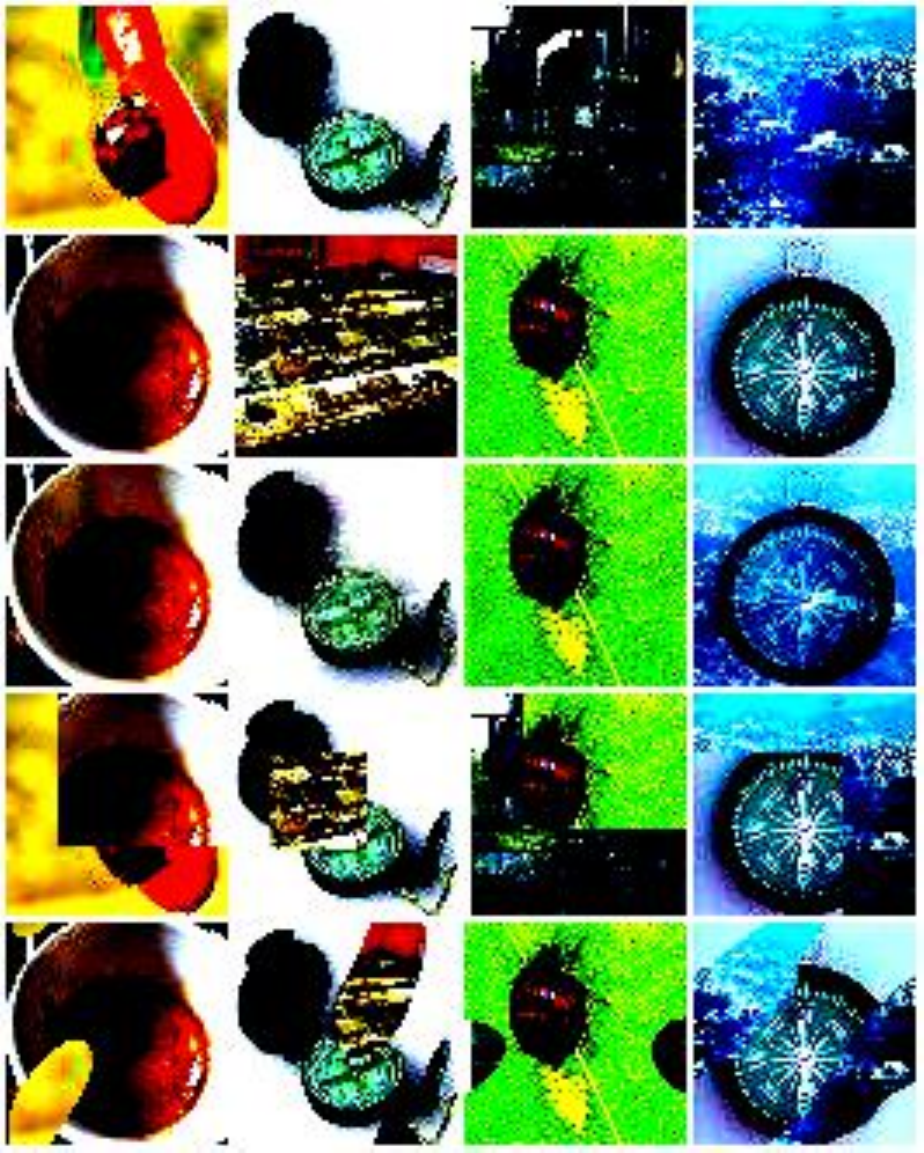} }\\[3em]
    Image 2 & \\[3em]
    \mixup & \\[3em]
    \cutmix & \\[3em]
    \fmix & \\[3em]
 
\end{tabularx}
    \caption{New samples generated by the mixed-augmentations we use in our analyses. }
    \label{fig:aug_ex}
\end{figure}

Figure~\ref{fig:aug_ex} provides examples of new samples obtained with \mixup, \cutmix, and \fmix. For each pair of images, we sample a mixing coefficient $\lambda \thicksim$ Beta(1, 1) for all three augmentations. 
For a large value of $\lambda$, as is the case in the first column, it can be seen that \cutmix covers less of Image 1 one than \fmix. As mentioned in the main body of the paper, this is because \cutmix allows the masking patch to lay outside of the image.

\section{Experimental details} \label{sup:exp_details}

Throughout the paper, we use PreAct-ResNet18~\citep{he2016identity} models, trained for 200 epochs with a batch size of 128.
For the MSDA parameters we use the same values as~\citet{harris2020understanding}.
All models are augmented with random crop and horizontal flip and are averaged across 5 runs.
We optimise using SGD with 0.9 momentum, learning rate of 0.1 up until epoch 100 and 0.001 for the rest of the training. 
This is due to an incompatibility with newer versions of the PyTorch library of the official implementation of~\citet{harris2020understanding}, which we use as a starting point. 
However, the difference in learning rate schedule between our work and prior art does not affect our findings since we are not introducing a new method to be applied at training time.
In our case, it is sufficient to show that the bias exists in at least one configuration. 
The models were trained on either one of the following: Titan X Pascal, GeForce GTX 1080ti or Tesla V100. 
For the analysis, a GeForce GTX 1050 was also used.

\subsection*{Training models}
The code for model training is largely based on the open-source official implementation of \fmix, which also includes those of \mixup, \cutout, and \cutmix.
For the experiment where we use the reformulated objective to combine data sets, instead of mixing with a permutation of the batch, as it is done in the original implementation of the mixed-augmentations, we now draw a batch form the desired data set. 
To ensure a fair comparison, for the basic we also perform inter-batch mixing.

\subsection*{Evaluating robustness}

For the \cutocc measurement, we modify open-source code to restrict the occluding patch to lie withing the the margins of the image to be occluded. This is to ensure that the mixing factor $\lambda$ matches the true proportion of the occlusion.
For \iocc, the implementation of Grad-CAM is again adapted from publicly available code.
With both methods, we evaluate 5 instances of the same model and average over the results obtained.

The added computation time of \iocc over the regular \cutocc for a fixed occlusion fraction is that of performing Grad-CAM on train and test data, as well as evaluating on the latter.
With a batch size of 128, this takes under half an hour. 

\section{Analysis of wrong predictions} \label{sup:wrongPred}

\begin{figure*}
    \begin{subfigure}[t]{0.49\linewidth}
        \centering
        \includegraphics[width=\linewidth]{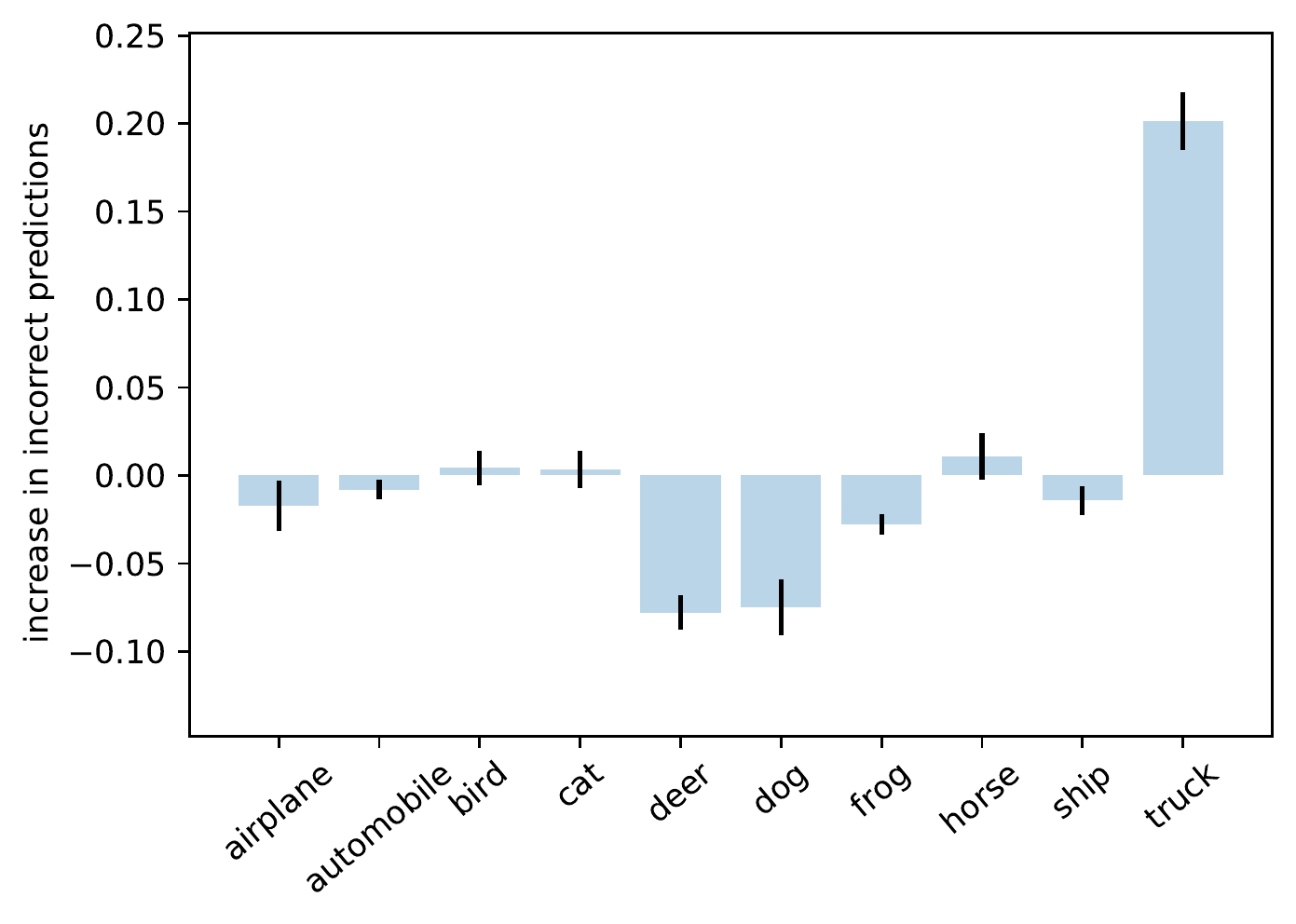}
        \caption{basic}
    \end{subfigure}
    \hfill 
    \begin{subfigure}[t]{0.49\linewidth} 
        \centering
        \includegraphics[width=\linewidth]{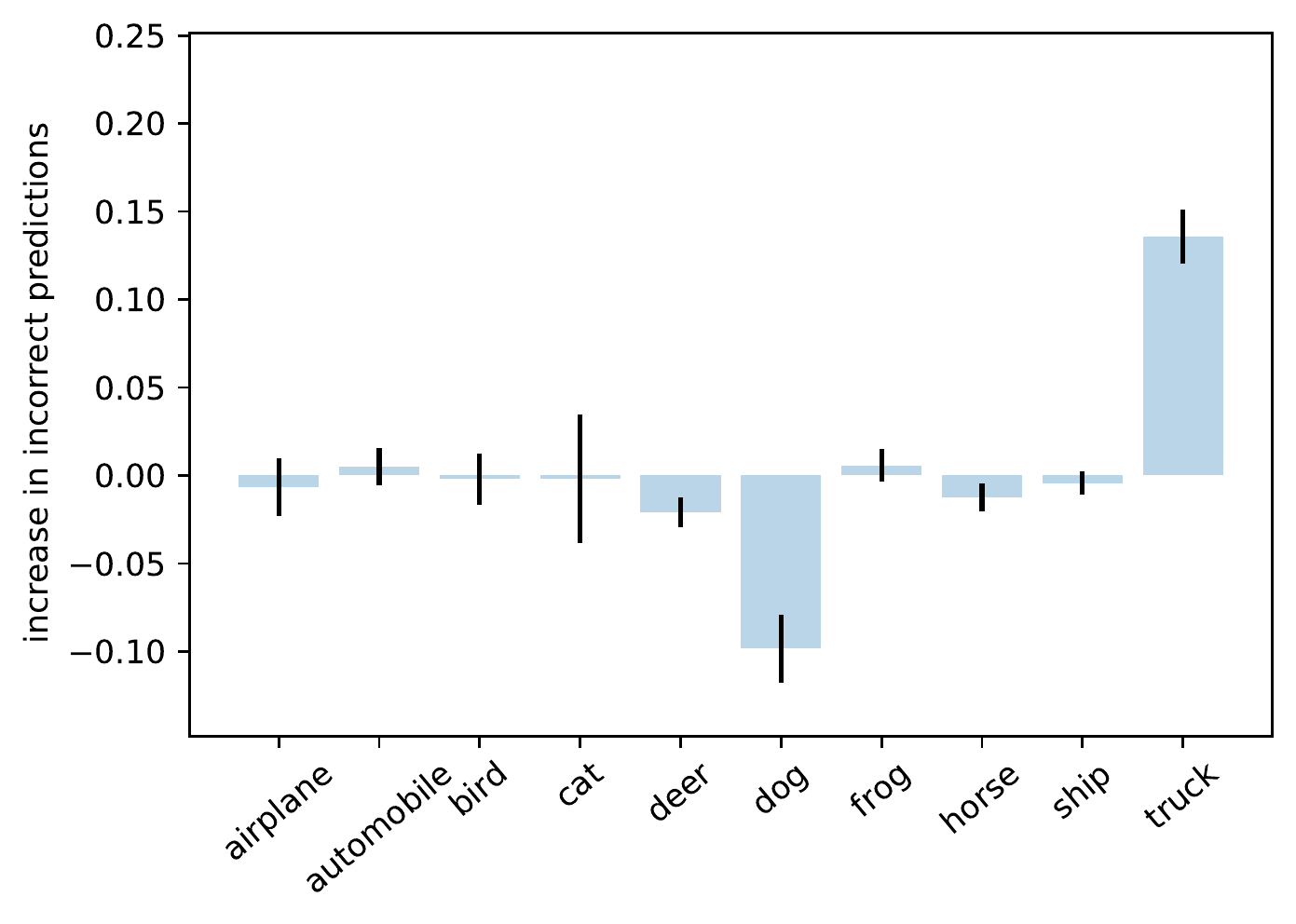}
        \caption{\mixup} 
    \end{subfigure}
    \newline
    \begin{subfigure}[t]{0.49\linewidth} 
        \centering
        \includegraphics[width=\linewidth]{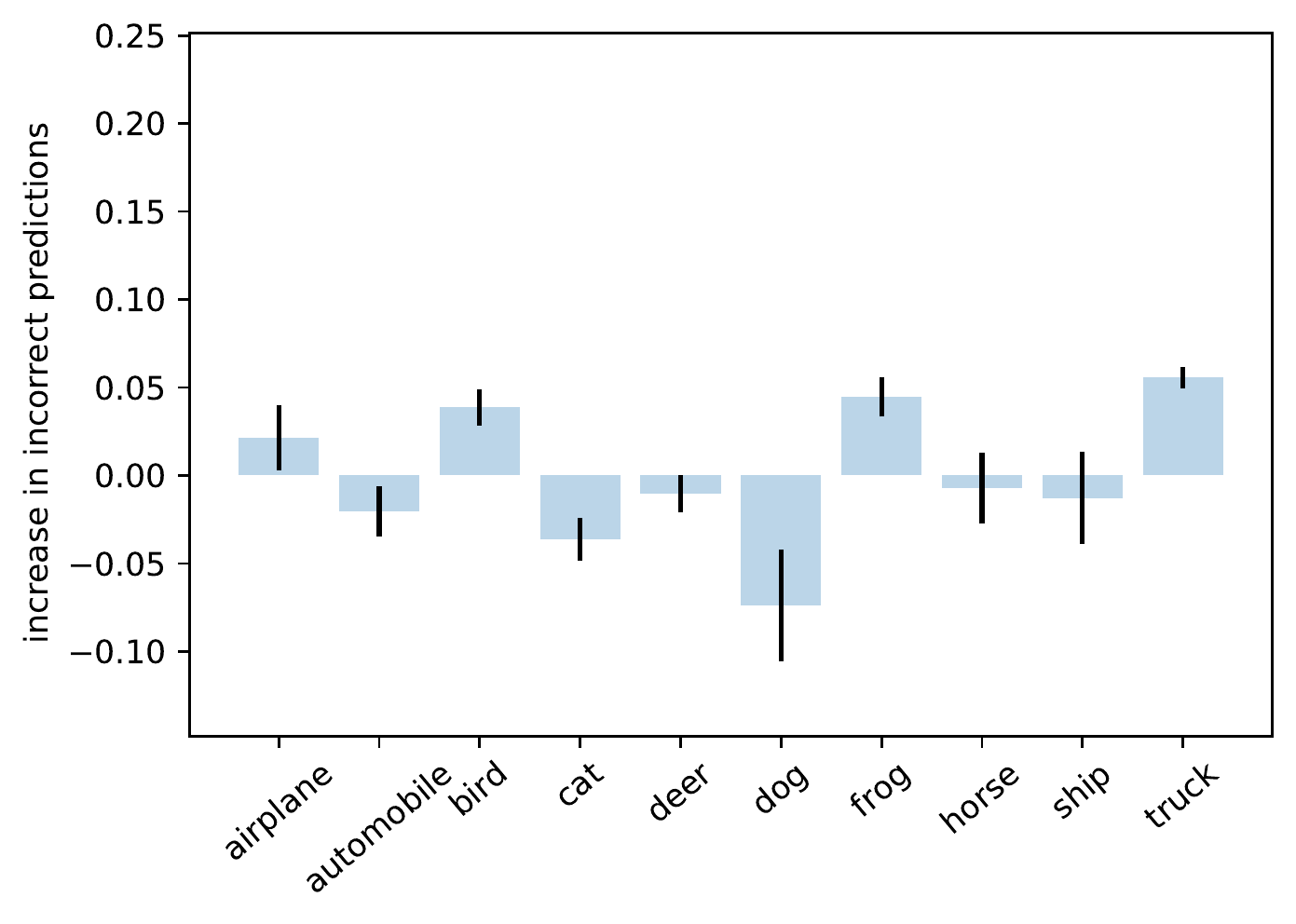}
        \caption{\cutmix} 
    \end{subfigure}
    \hfill 
    \begin{subfigure}[t]{0.49\linewidth} 
        \centering
        \includegraphics[width=\linewidth]{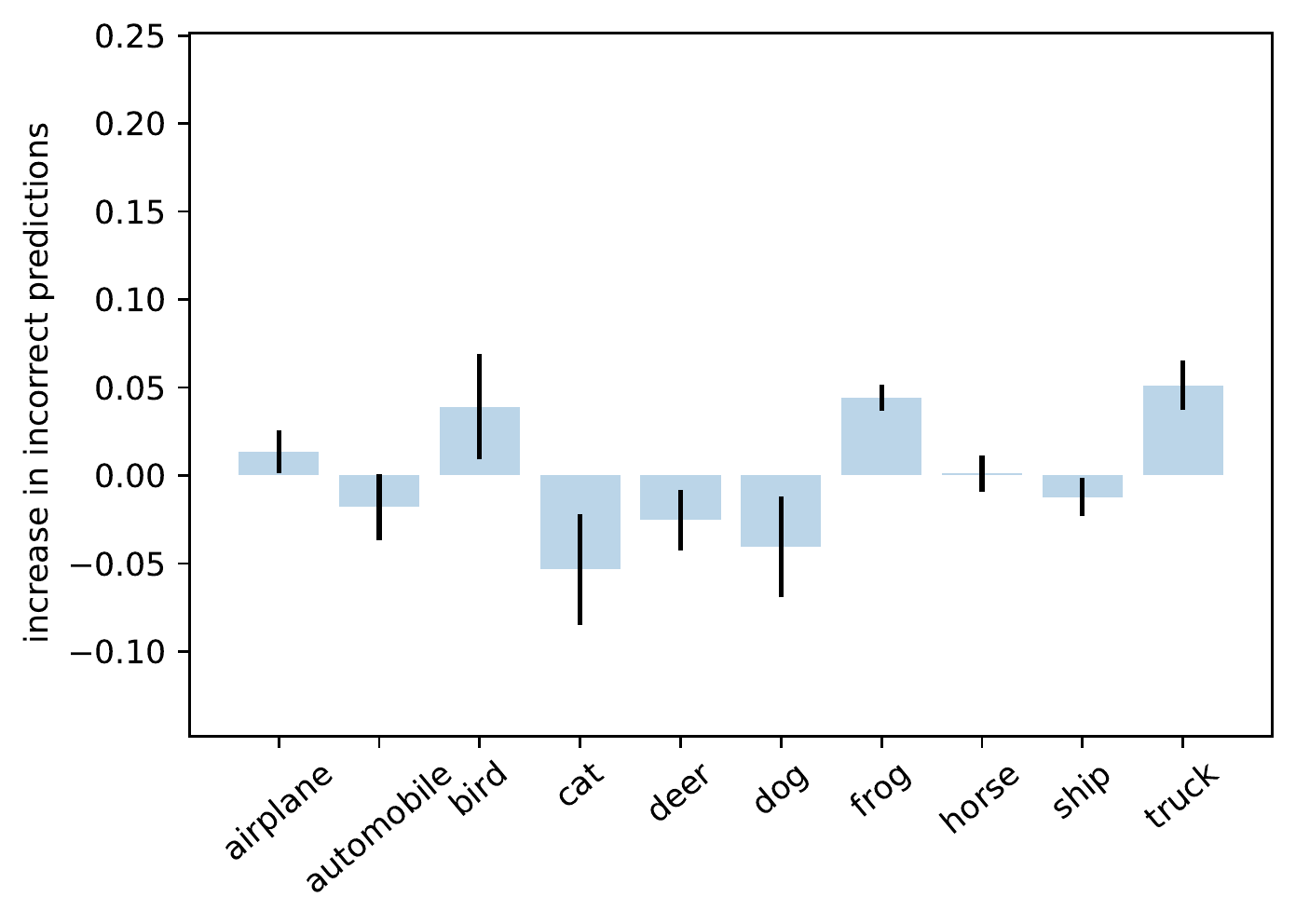}
        \caption{\fmix} 
    \end{subfigure}
    \caption{Difference between wrongly predicted classes when testing on original data versus \cutmix images. The evaluated models from left to right, top to bottom are trained on \cifar{10} with: no mixed-data augmentation (basic), \mixup, \cutmix, and \fmix.}
    \label{fig:wrong_pred_cutmix_c10}
\end{figure*}

\begin{figure*}
    \begin{subfigure}[t]{0.49\linewidth}
        \centering
        \includegraphics[width=\linewidth]{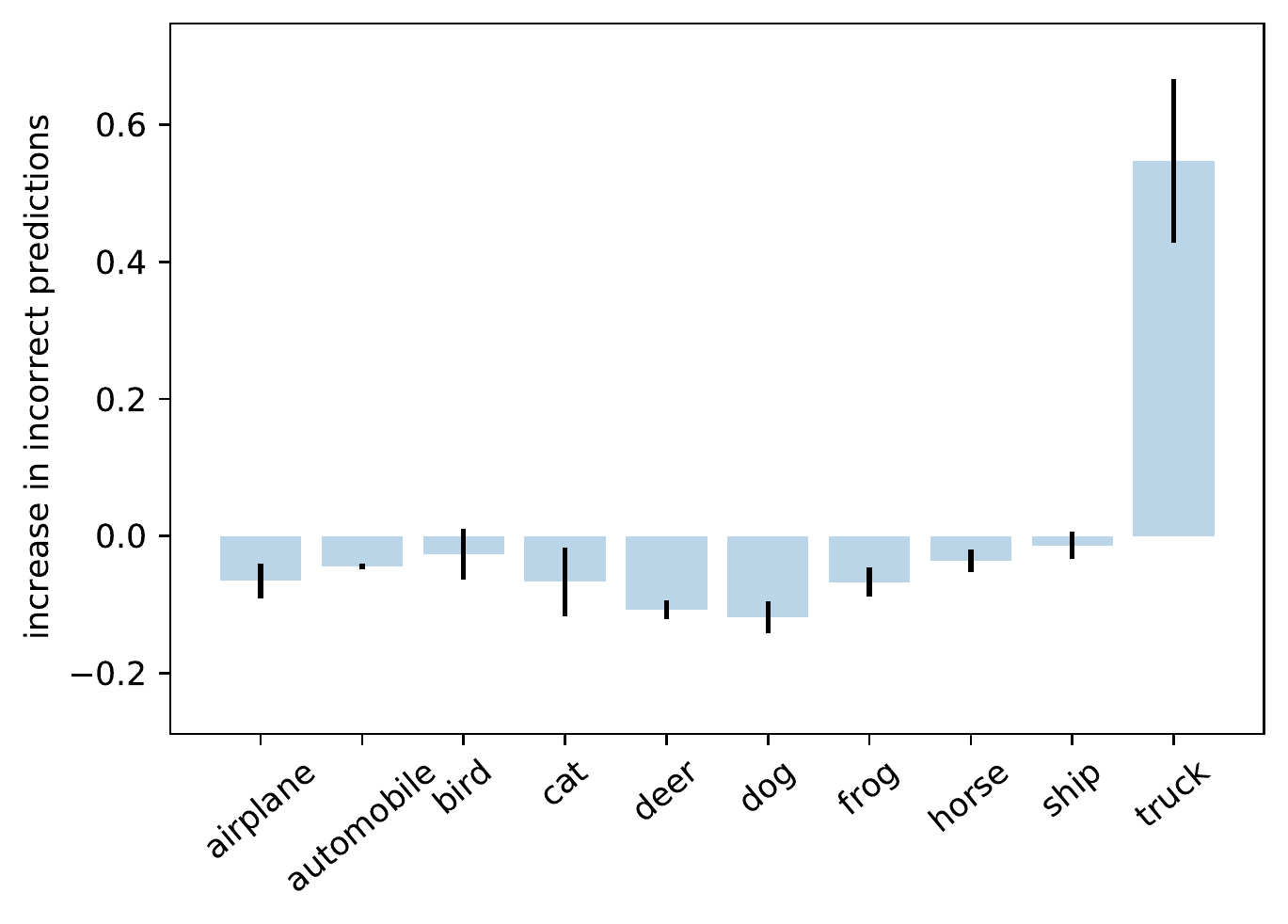}
        \caption{basic}
    \end{subfigure}
    \hfill 
    \begin{subfigure}[t]{0.49\linewidth} 
        \centering
        \includegraphics[width=\linewidth]{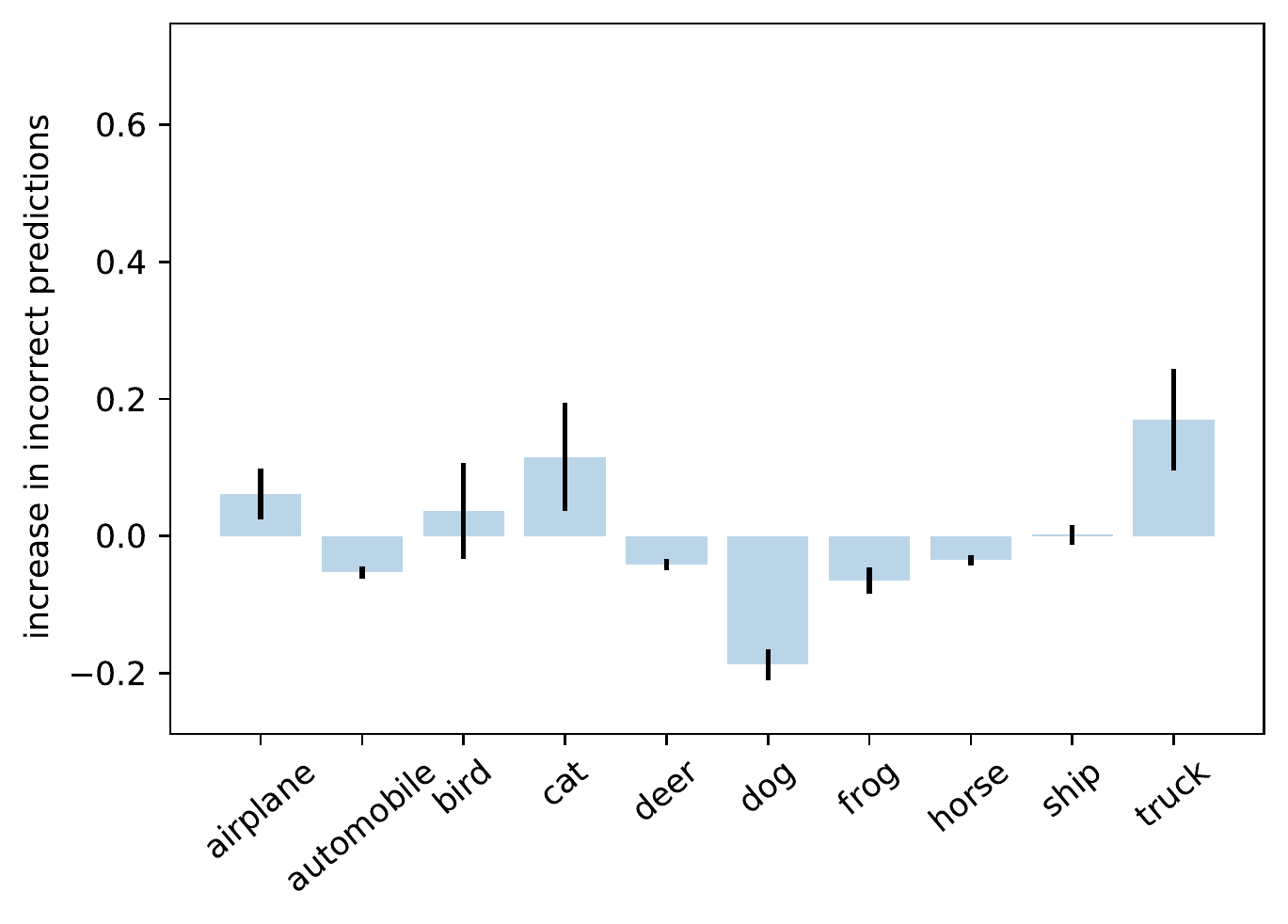}
        \caption{\mixup} 
    \end{subfigure}
    \newline
    \begin{subfigure}[t]{0.49\linewidth} 
        \centering
        \includegraphics[width=\linewidth]{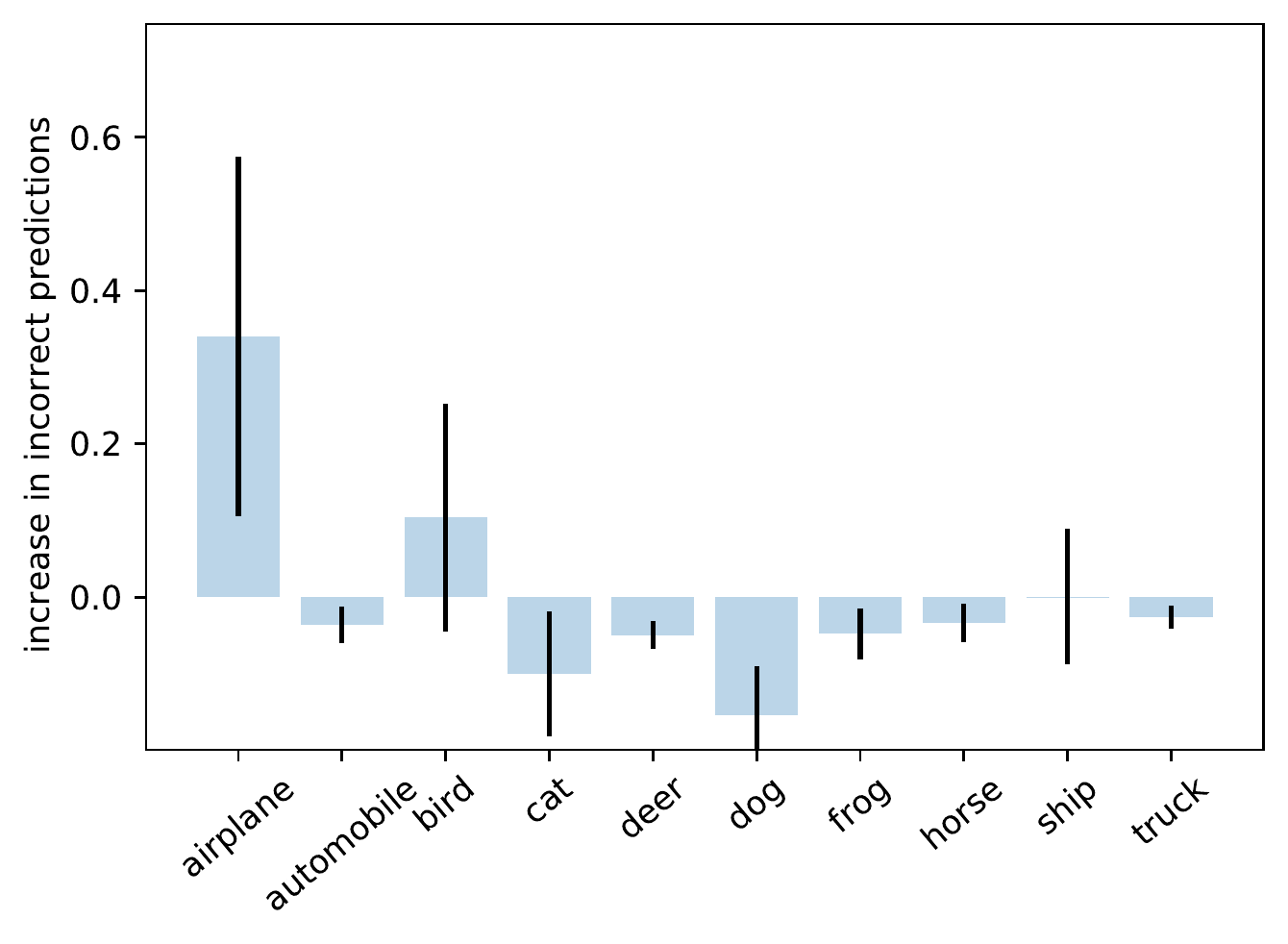}
        \caption{\cutmix} 
    \end{subfigure}
    \hfill 
    \begin{subfigure}[t]{0.49\linewidth} 
        \centering
        \includegraphics[width=\linewidth]{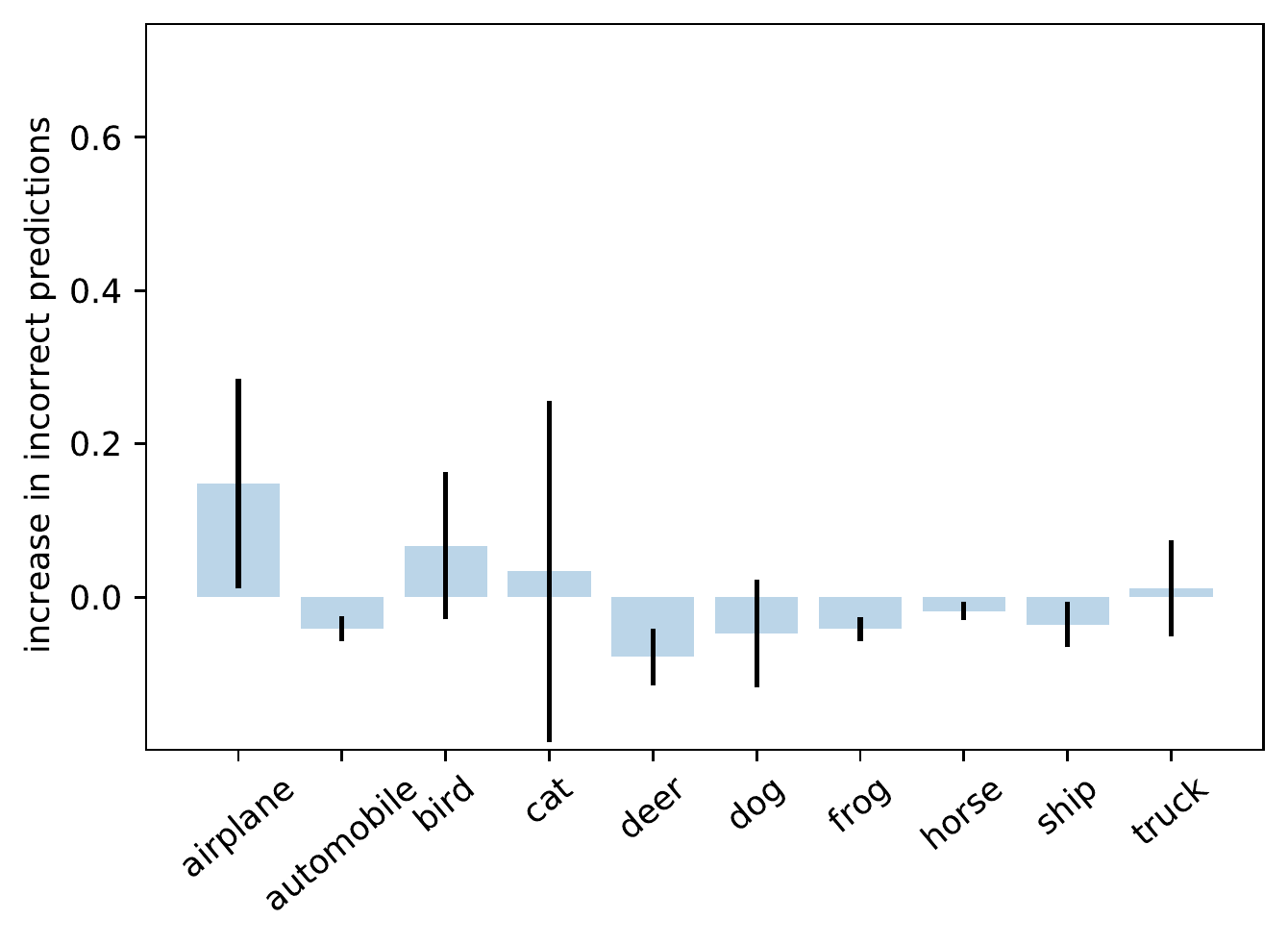}
        \caption{\fmix} 
    \end{subfigure}
    \caption{Difference between wrongly predicted classes when testing on original data versus \cutout{} images. The evaluated models from left to right, top to bottom are trained on \cifar{10} with: no mixed-data augmentation (basic), \mixup, \cutmix, and \fmix.}
    \label{fig:wrong_pred_cutout_c10}
\end{figure*}

This section provides visual results of the experiments carried out to identify whether \cutocc provides biased results.
We reintroduce the experiment here. 
For each class, when presented with regular test data, we count the number of times a sample was incorrectly identified as belonging to that class.
We subtract the obtained values from the number of misidentifications on distorted data.
Figure~\ref{fig:wrong_pred_cutout_c10} shows results for the \cifar{10} data set when occluding with uniform patches, while Figure~\ref{fig:wrong_pred_cutmix_c10} show results for the same problem but where occluding is done with patches corresponding to images from the \cifar{100} data set.
We mix with images from a different data set to avoid issues arising from an image belonging to two classes simultaneously.
In Section~\ref{sup:archi}, we also show results for more data sets and architectures, while Section~\ref{sec:3msks_bias} contains the increase in wrong predictions obtained for the RM and RM3 models.

\section{Training with fixed random masks} \label{sup:3msks}

\begin{figure*} 
    \begin{subfigure}[t]{0.49\linewidth}
        \centering
        \includegraphics[width=\linewidth]{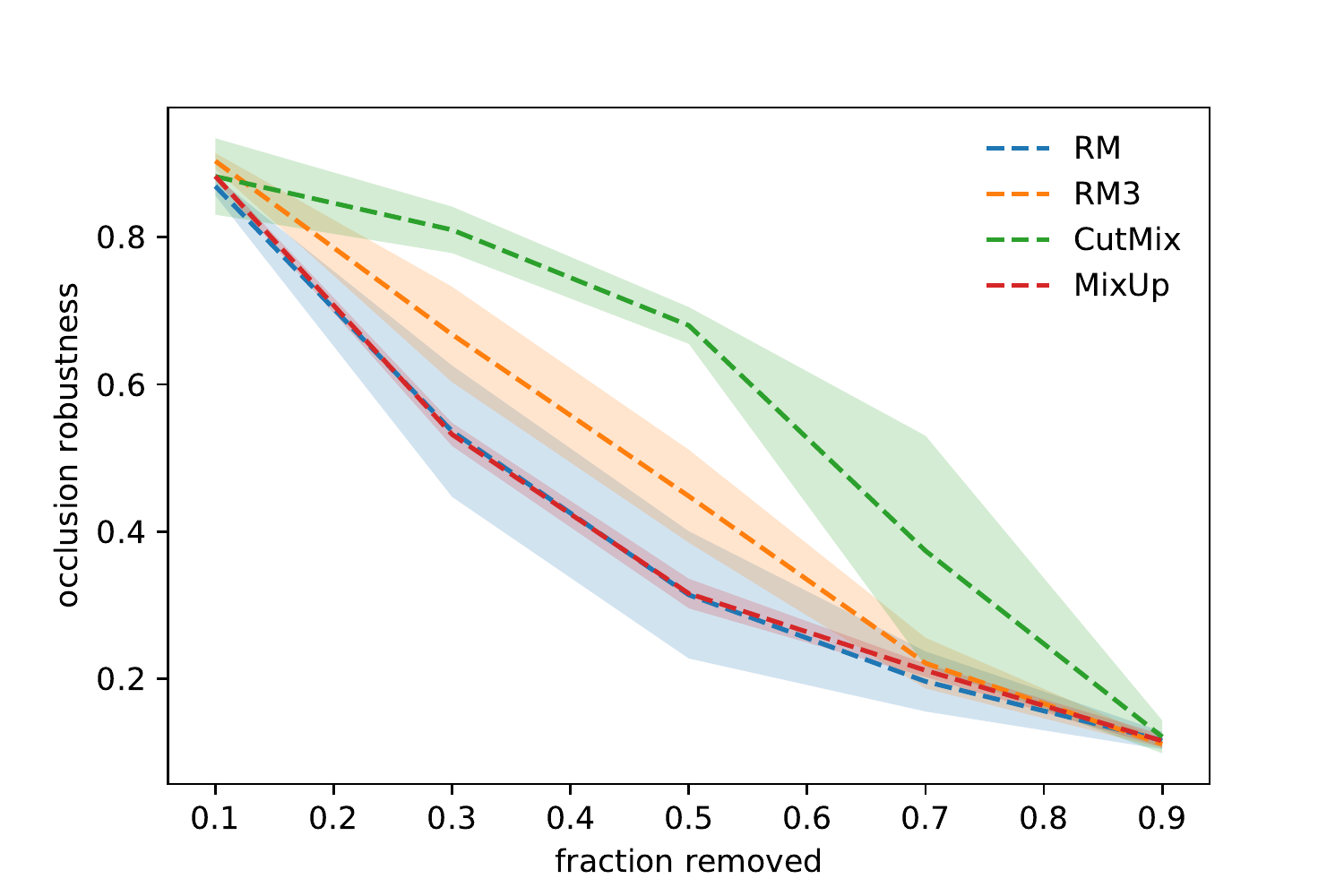}
    \end{subfigure}
    \hfill 
    \begin{subfigure}[t]{0.49\linewidth}
        \centering
        \includegraphics[width=\linewidth]{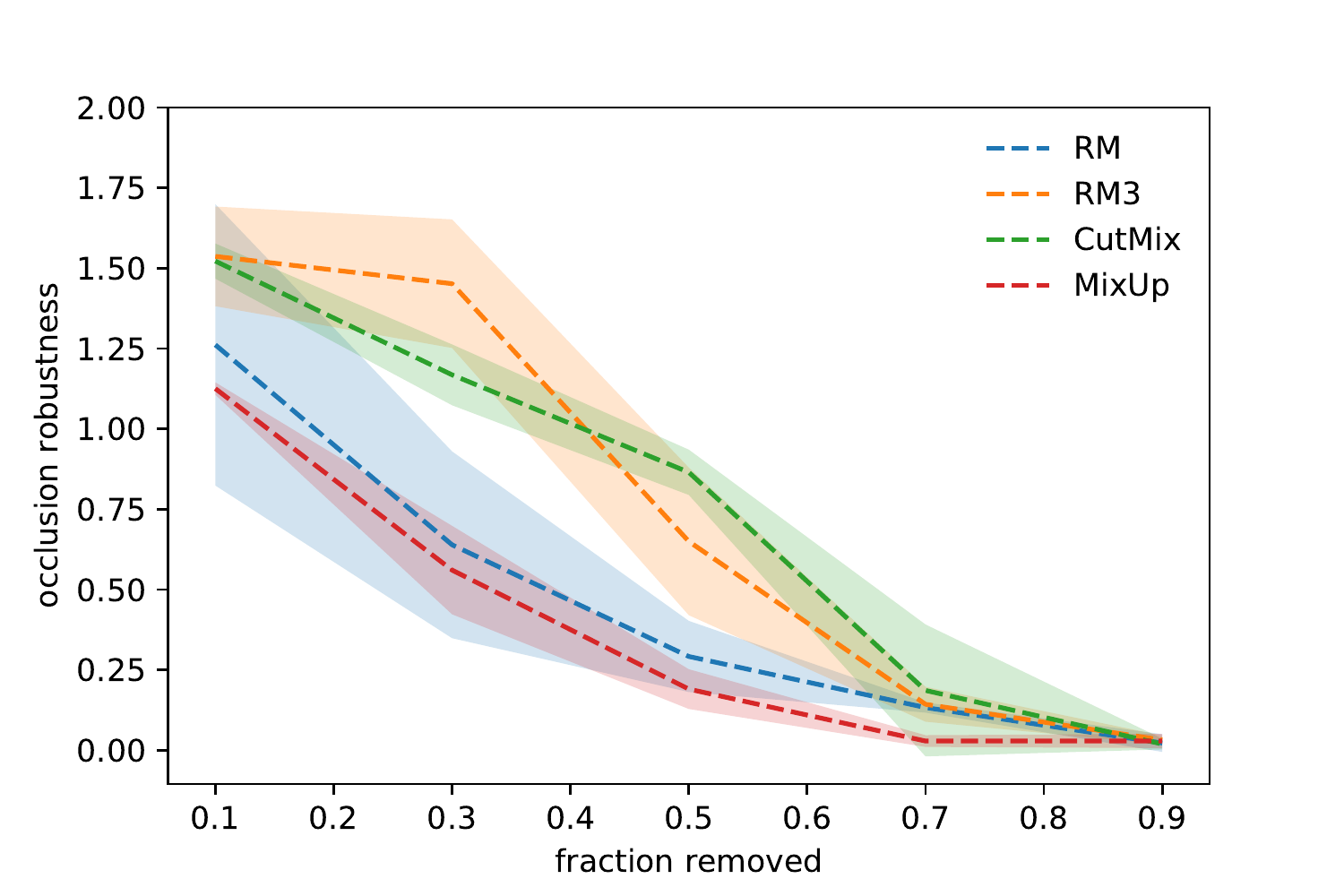}
    \end{subfigure}
    \caption{ \cutocc (left) and \iocc (right).
    }
    \label{fig:3msks}
\end{figure*}

Figure~\ref{fig:3msks} gives the results for \cutocc and \iocc for training with 1 or 3 random masks sampled from Fourier space. We provide \cutmix and \mixup as references and exclude the basic and \fmix for visual clarity.

\section{Occluding with images from another data set} \label{sup:mix_nomix}

\begin{figure*}
    \begin{subfigure}[t]{0.49\linewidth}
        \centering
        \includegraphics[width=\linewidth]{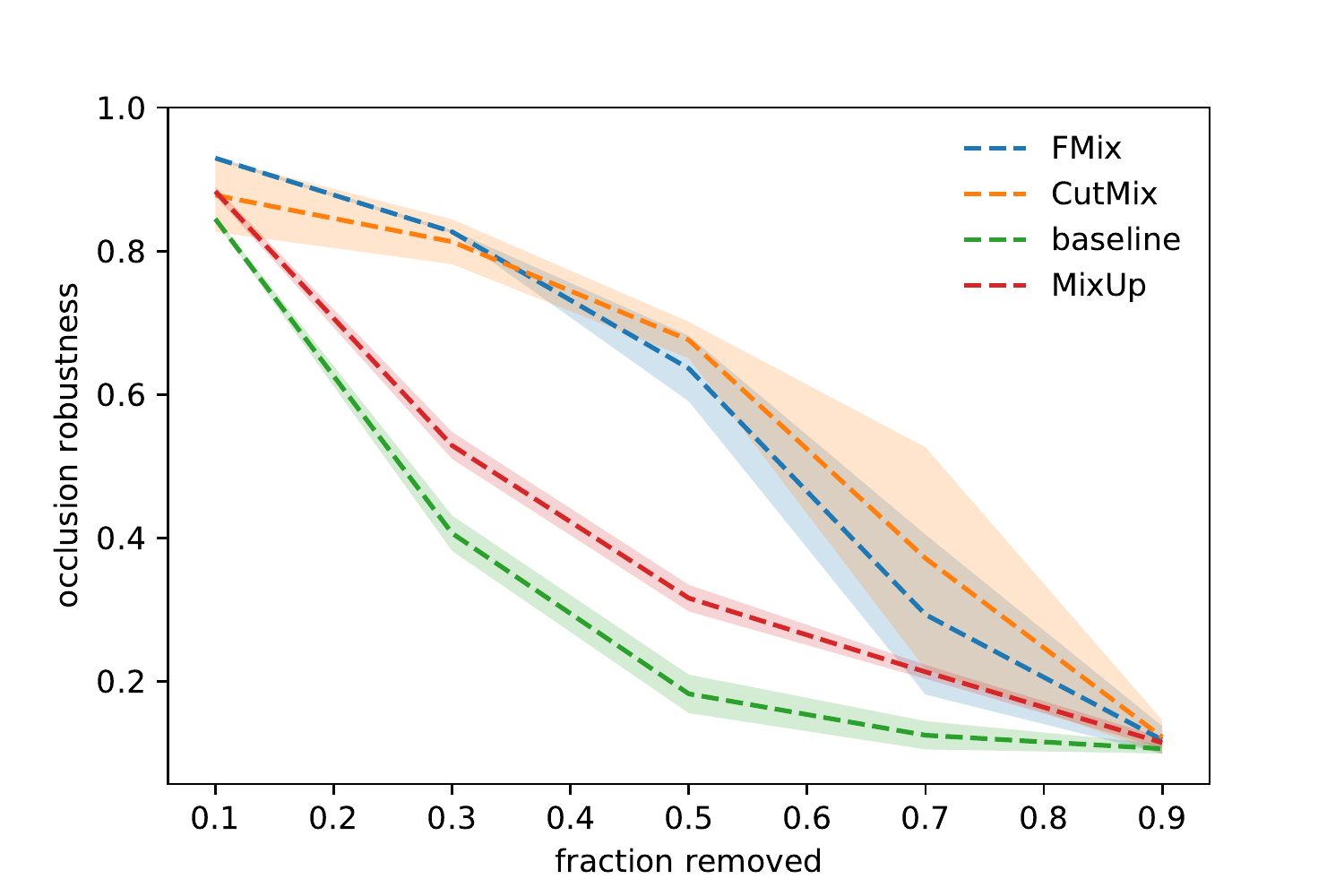}
        \caption{Regular \cutocc}
    \end{subfigure}
    \hfill 
    \begin{subfigure}[t]{0.49\linewidth} 
        \centering
        \includegraphics[width=\linewidth]{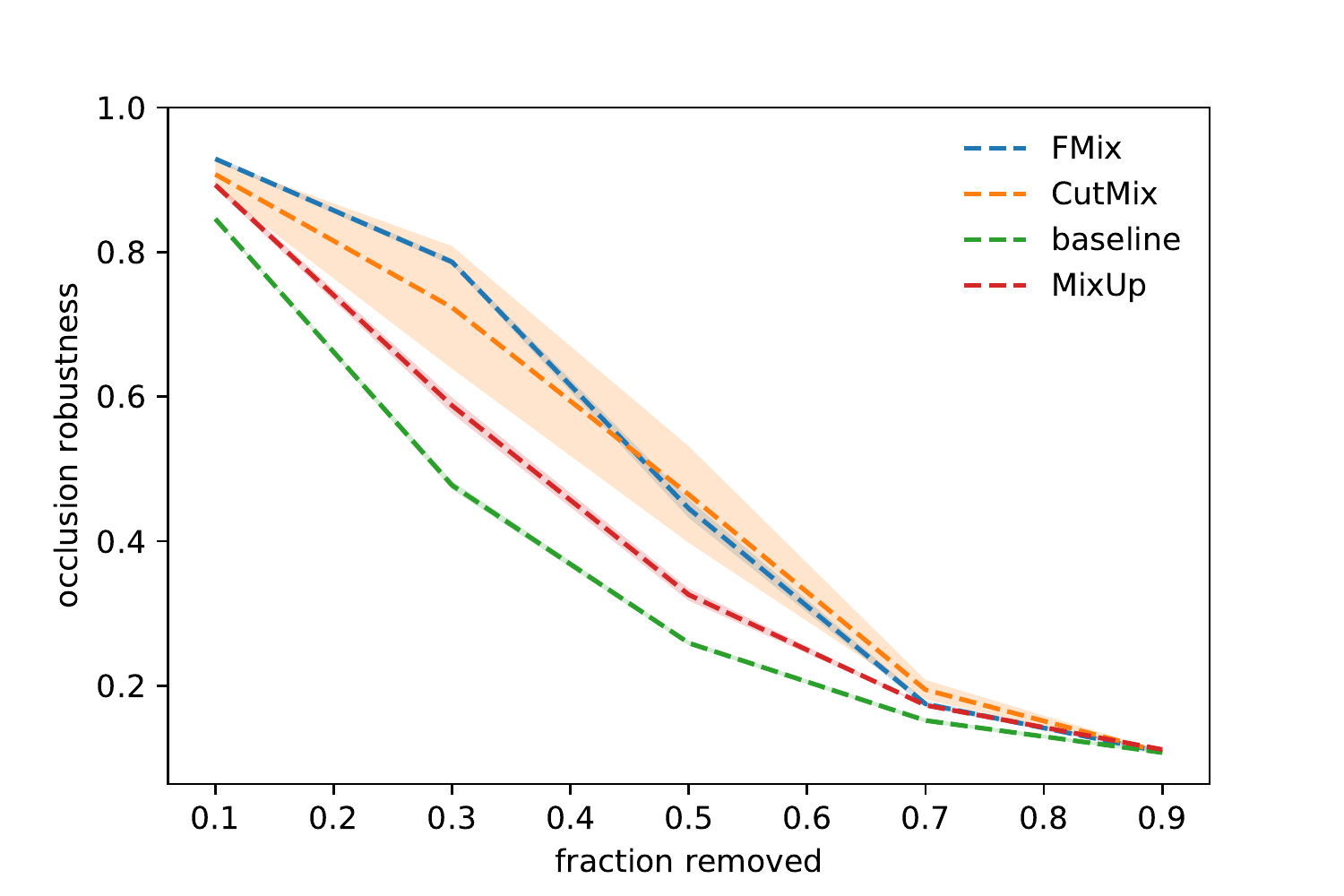}
        \caption{Mixing \cutocc} 
    \end{subfigure}
    \newline
    \begin{subfigure}[t]{0.49\linewidth} 
        \centering
        \includegraphics[width=\linewidth]{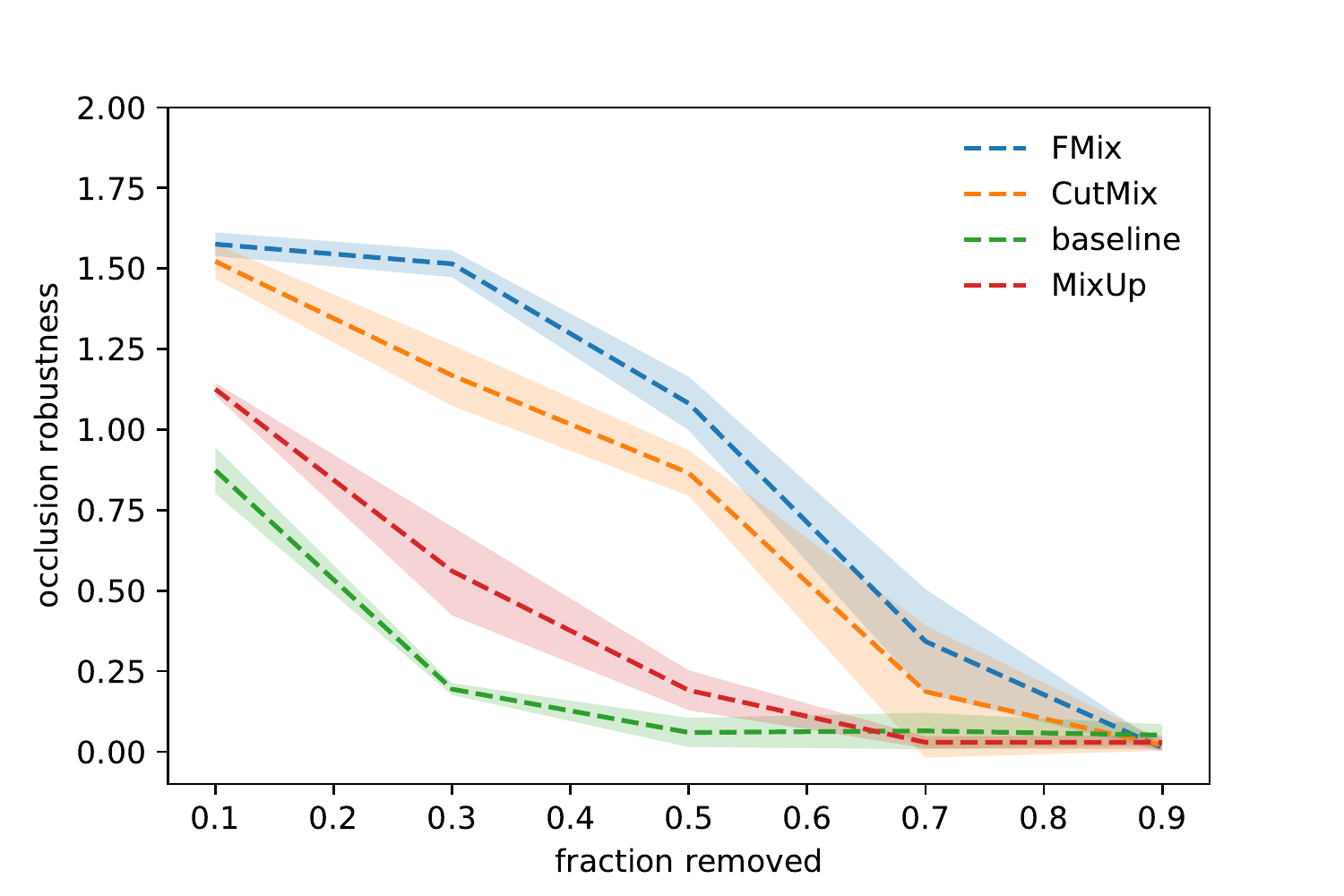}
        \caption{Regular \iocc} 
    \end{subfigure}
    \hfill 
    \begin{subfigure}[t]{0.49\linewidth} 
        \centering
        \includegraphics[width=\linewidth]{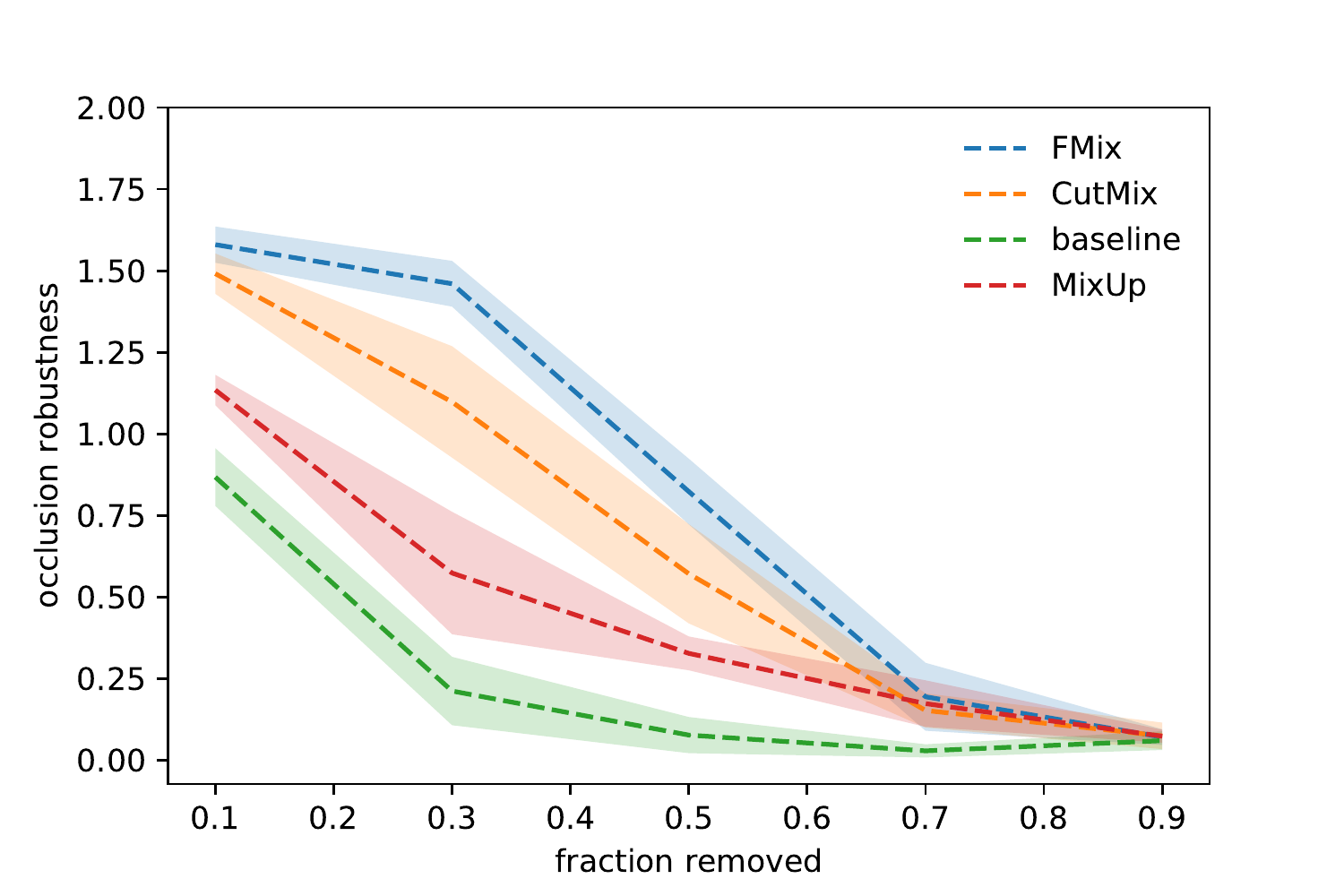}
        \caption{Mixing  \iocc} 
    \end{subfigure}
    \caption{Comparison of metric sensitivity to textured occlusion. Regular occlusion refers to superimposing uniform patches over \cifar{10} images, while mixing refers to superimposing part of \cifar{100} samples. Mixing \cutocc provides significantly different results to its regular counterpart.}
    \label{fig:mix_nomix}
\end{figure*}

Since \cutocc does not account for the bias introduced by the occluding method, it is expected that changing the patch to a non-uniform one would greatly affect the results. For \cifar{10} models, Figure~\ref{fig:mix_nomix} presents the results of occluding with \cifar{100} images. 

\section{Randomising labels} \label{sup:rand_lbls}

\begin{figure}
    \centering
    \includegraphics[width=\linewidth]{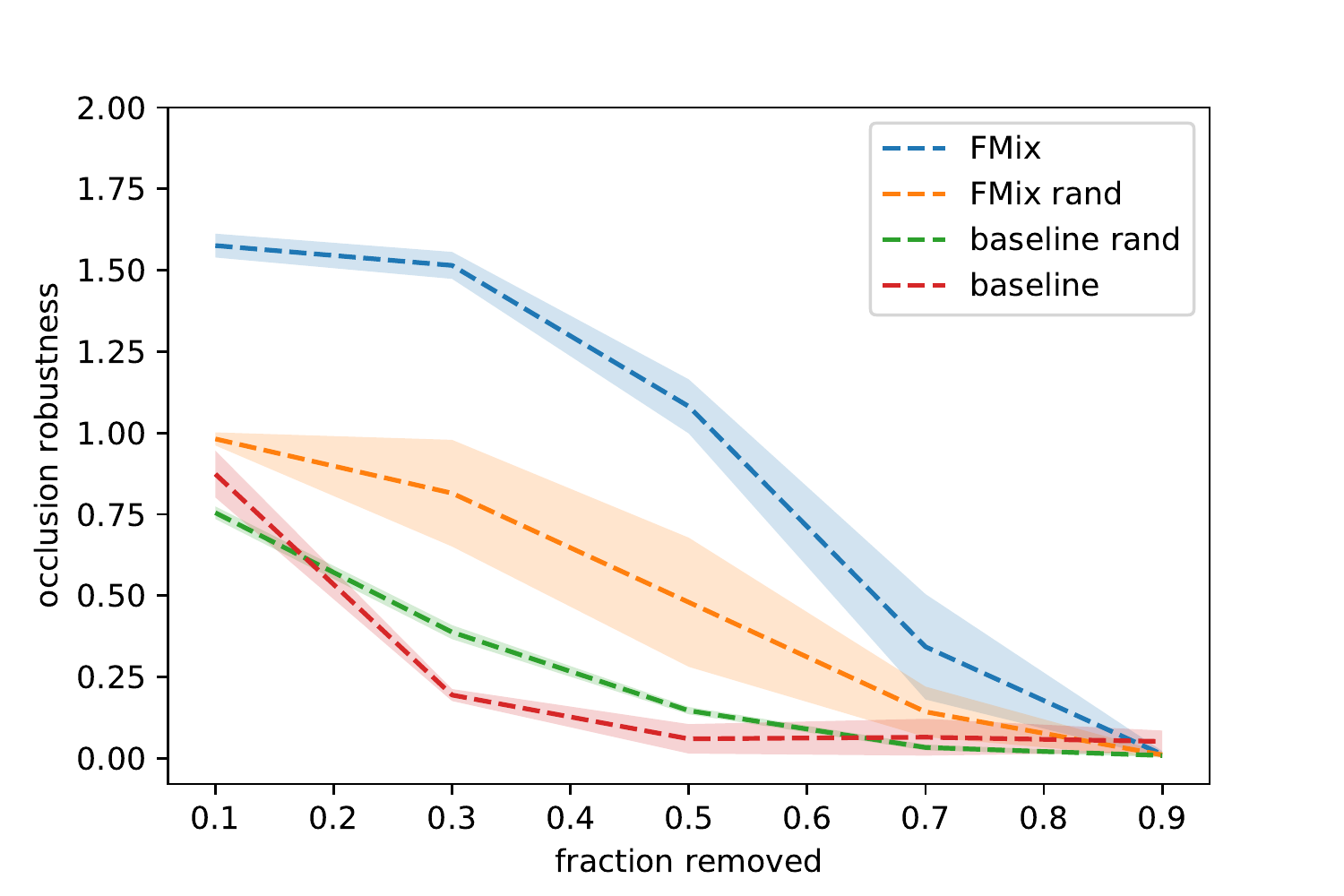}
    \caption{\iocc results for training with clean and corrupted labels for basic and \fmix augmentation.} 
    \label{fig:rand}
\end{figure}

To assess the sensitivity of \cutocc and \iocc to the overall performance of the model, we also experiment with randomising all the labels of the \cifar{10} data set.
When evaluated on the unaugmented training data, all the basic models achieve 100\% accuracy, while the \fmix models reach $99.99_{\pm0.01}$.
Figure \ref{fig:rand} gives the \iocc scores for these models.

\section{Distribution of wrong predictions for other data sets and architectures} \label{sup:archi} 

\begin{figure*}
    \centering
    \includegraphics[width=\linewidth]{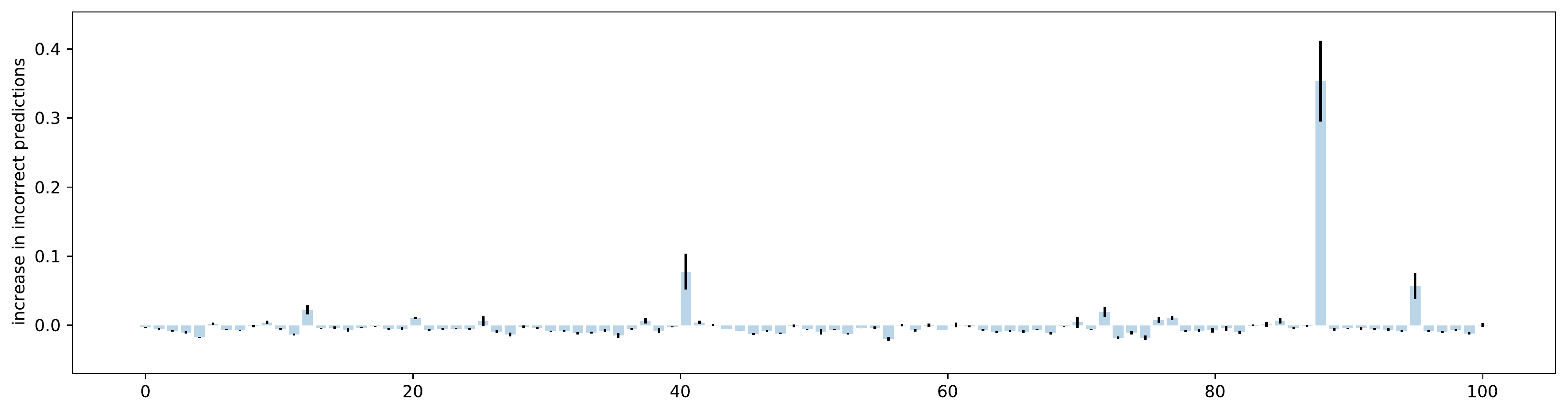}
    \caption{\cifar{100} results for the basic model  for \cutout images.} 
    \label{fig:fig:wrong_pred_cutout_c100}
\end{figure*}

\begin{figure*}
    \begin{subfigure}[t]{0.49\linewidth} 
        \centering
        \includegraphics[width=\linewidth]{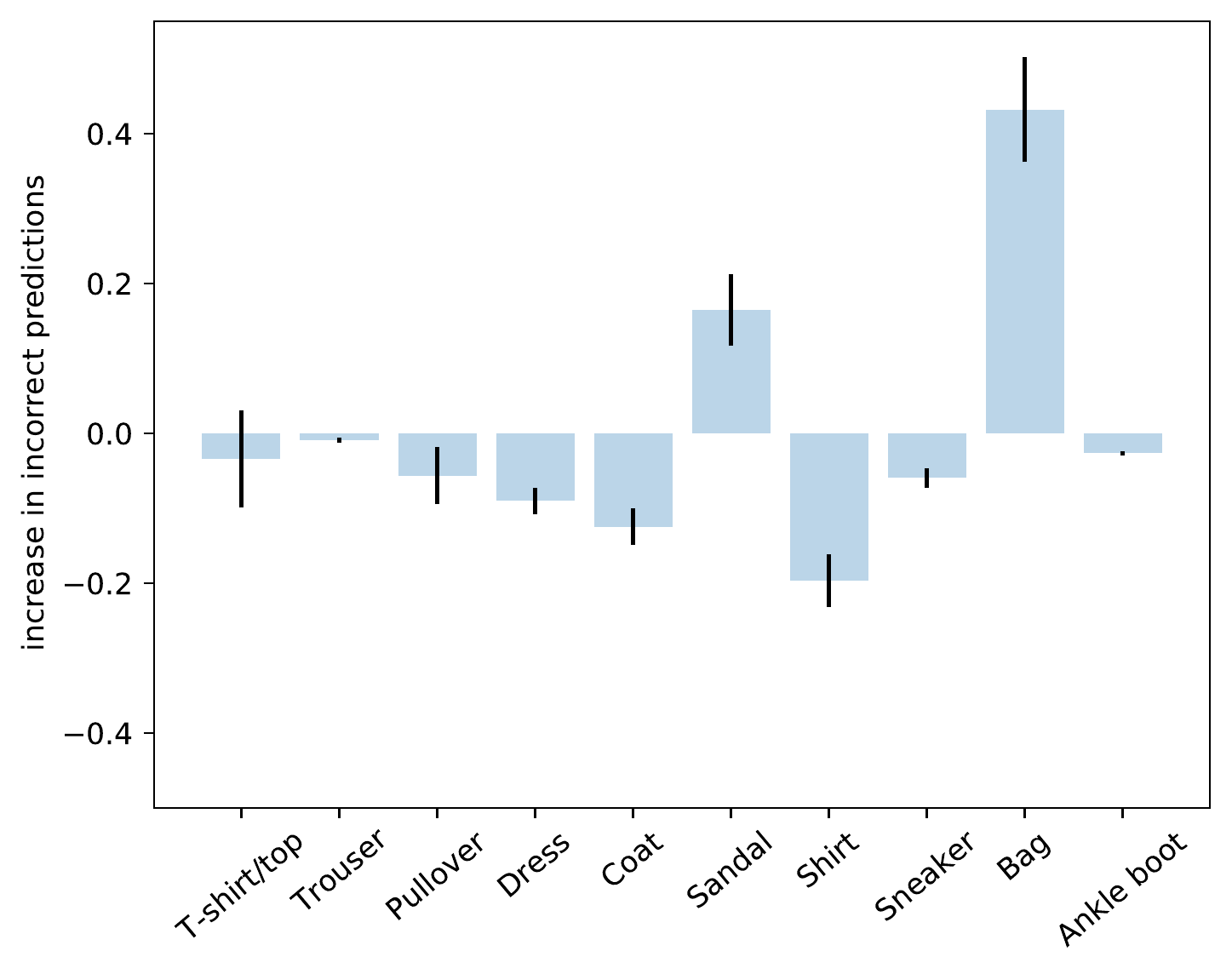}
        \caption{\mixup results on Fashion MNIST. Note that for this data set, when using uniform patches, only \mixup exhibits a visible bias.} 
    \label{fig:wrong_pred_cutout_fash}
    \end{subfigure}
    \hfill 
    \begin{subfigure}[t]{0.49\linewidth} 
        \centering
        \includegraphics[width=\linewidth]{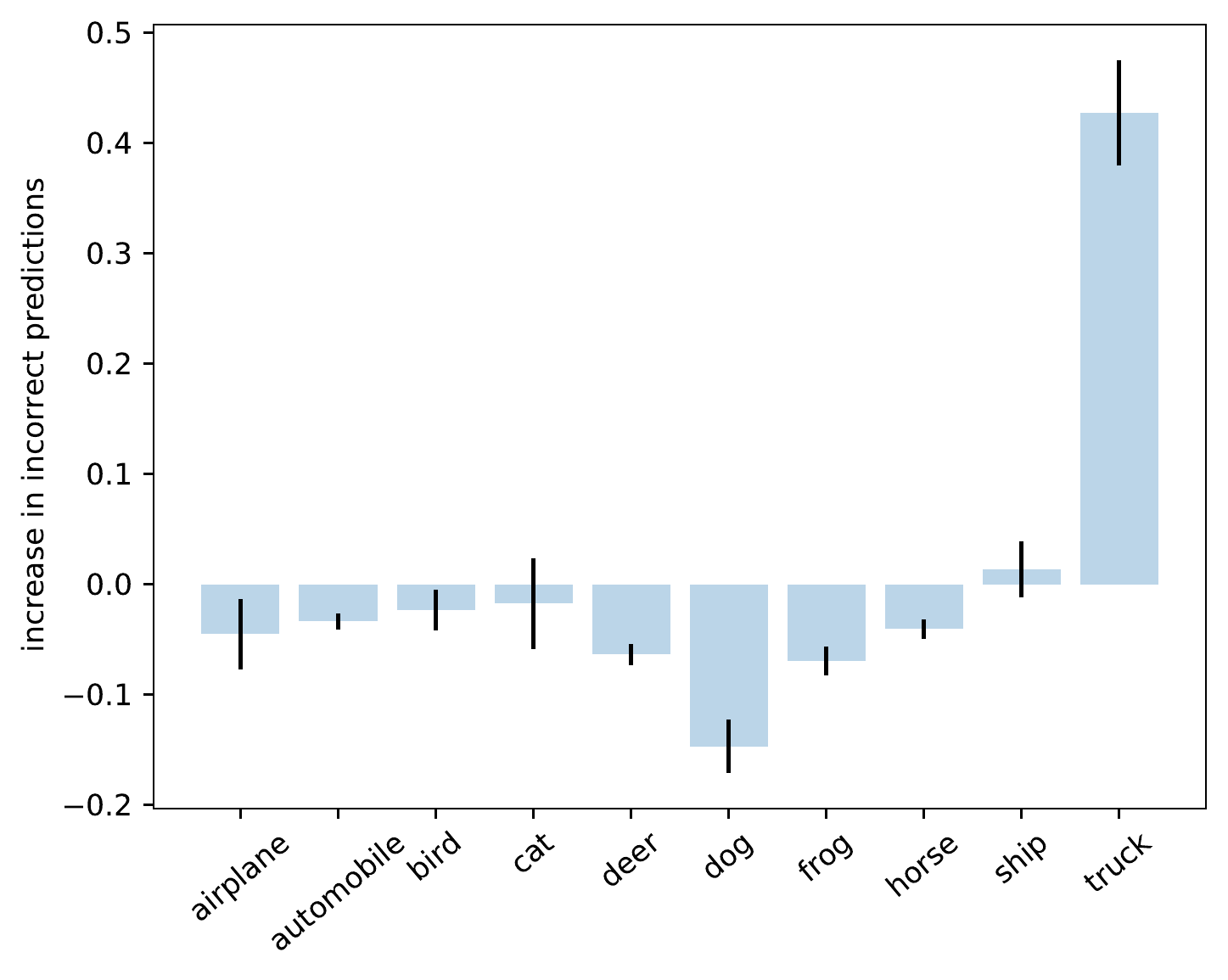}
        \caption{Results for the basic VGG-16 model on \cutout \cifar{10} images.}
        \label{fig:vgg}
    \end{subfigure}
    \caption{PreAct-ResNet18 on Fashion-MNIST (left) and VGG-16 on \cifar{10} (right) when occluding with black patches.}
    \label{fig:other}
\end{figure*}

In this section we include examples of identified bias for more data sets and architectures as follows: Figure~\ref{fig:fig:wrong_pred_cutout_c100} for \cifar{100}, Figure~\ref{fig:wrong_pred_cutout_fash} for Fashion MNIST~\citep{xiao2017fashion} and Figure~\ref{fig:vgg} for VGG-16~\citep{simonyan2014very} models.
In subsequent sections of the paper we also experiment with Tiny~\imagenet and \imagenet data sets, as well as BagNet models.

\section{\cifar{100} results for \cutocc and \iocc} \label{sup:c100Comp} 

\begin{figure*}
    \begin{subfigure}[t]{0.49\linewidth} 
        \centering
        \includegraphics[width=\linewidth]{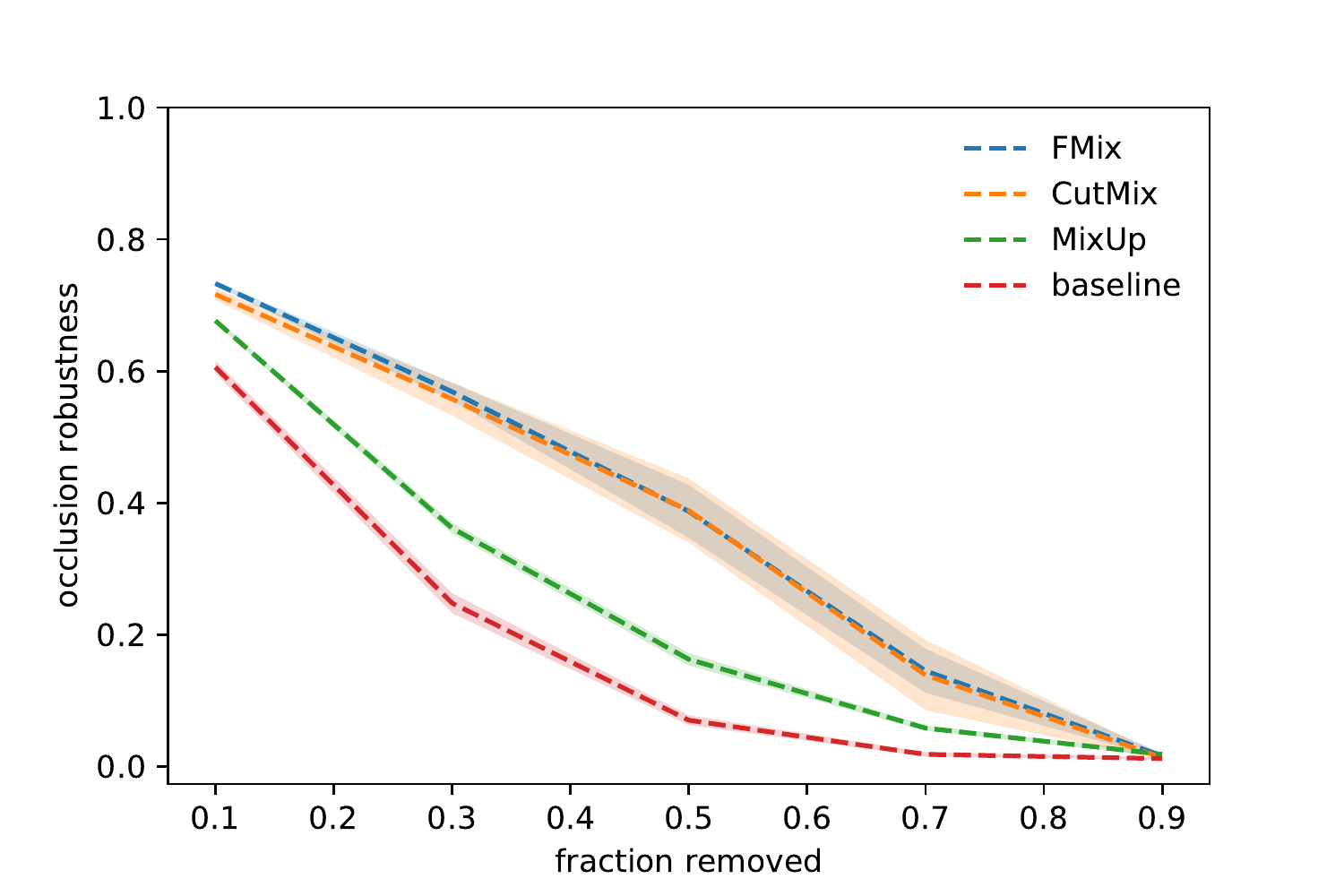}
    \end{subfigure}
    \hfill 
    \begin{subfigure}[t]{0.49\linewidth} 
        \centering
        \includegraphics[width=\linewidth]{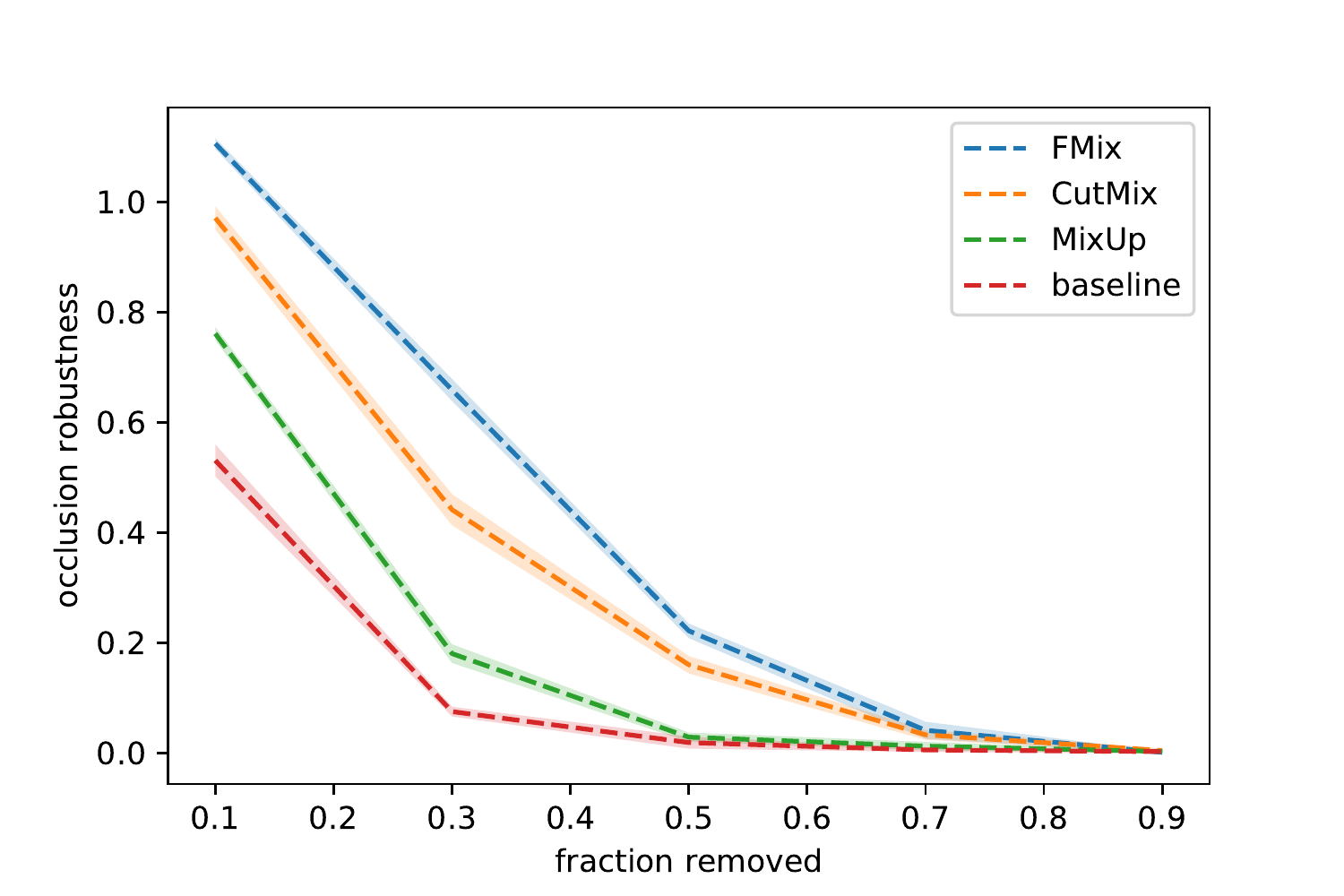}
    \end{subfigure}
    \caption{\cutocc(left) and \iocc(right) on the \cifar{100} data set. Note again that there is a difference in scale and for comparing the two methods we look at how they position the models with respect to each other.}
    \label{fig:c100_comparison}
\end{figure*}

Although, as stated in the main paper, we believe it is difficult to make a direct comparison between \cutocc and \iocc, Figure~\ref{fig:c100_comparison} gives the scores for the two methods on the \cifar{100} data set.
Again, \cutocc only assesses the robustness against contiguous patches, making no difference between \cutmix and \fmix.
More importantly, \iocc captures the more rapid fall of the masking methods, which in the case of \cutocc appear significantly more robust than basic due to their higher invariance to the artefacts introduced.

\section{Bias of fixed random masks} \label{sec:3msks_bias}

\begin{figure*}
    \begin{subfigure}[t]{0.49\linewidth} 
        \centering
        \includegraphics[width=\linewidth]{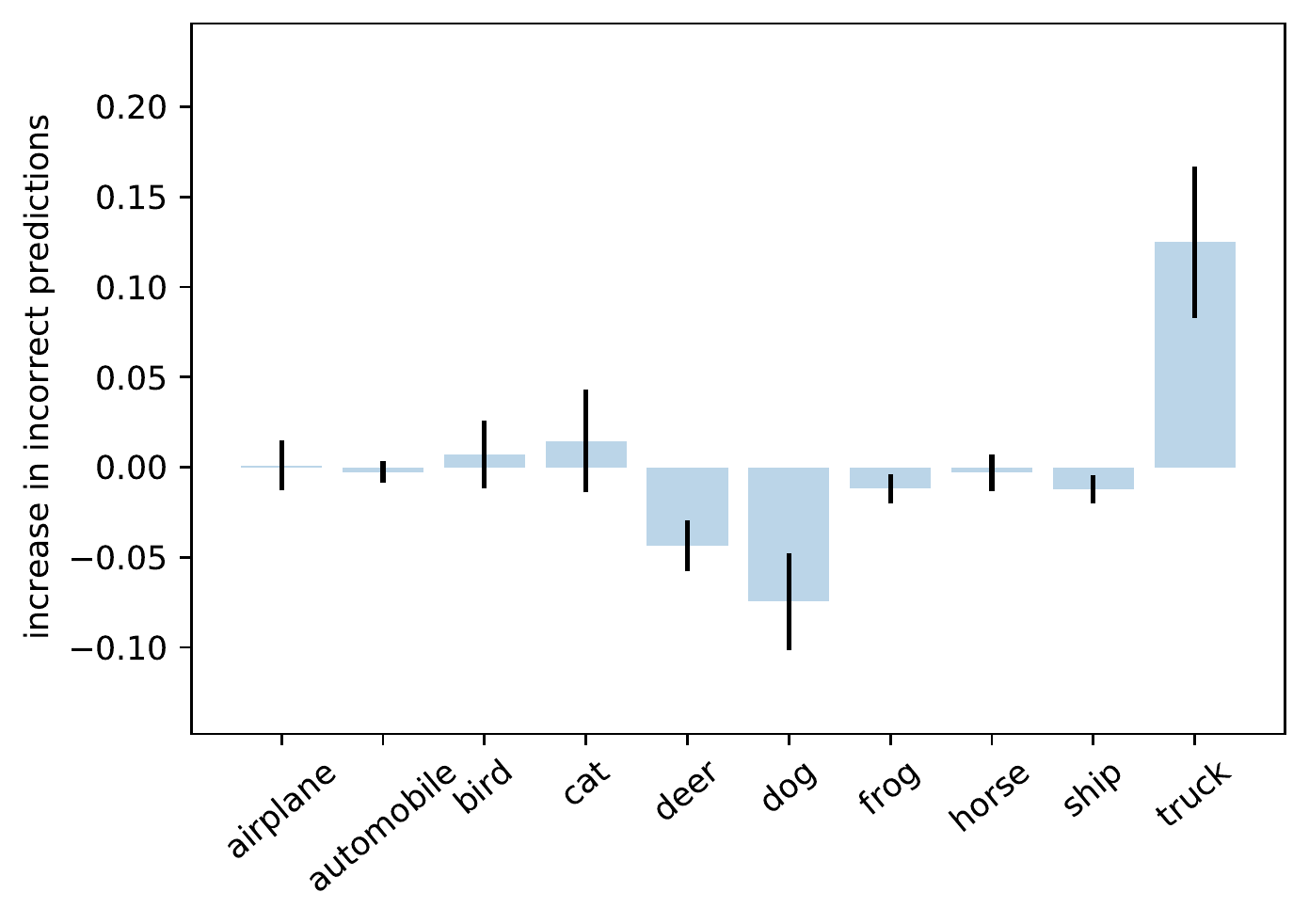}
    \end{subfigure}
    \hfill 
    \begin{subfigure}[t]{0.49\linewidth} 
        \centering
        \includegraphics[width=\linewidth]{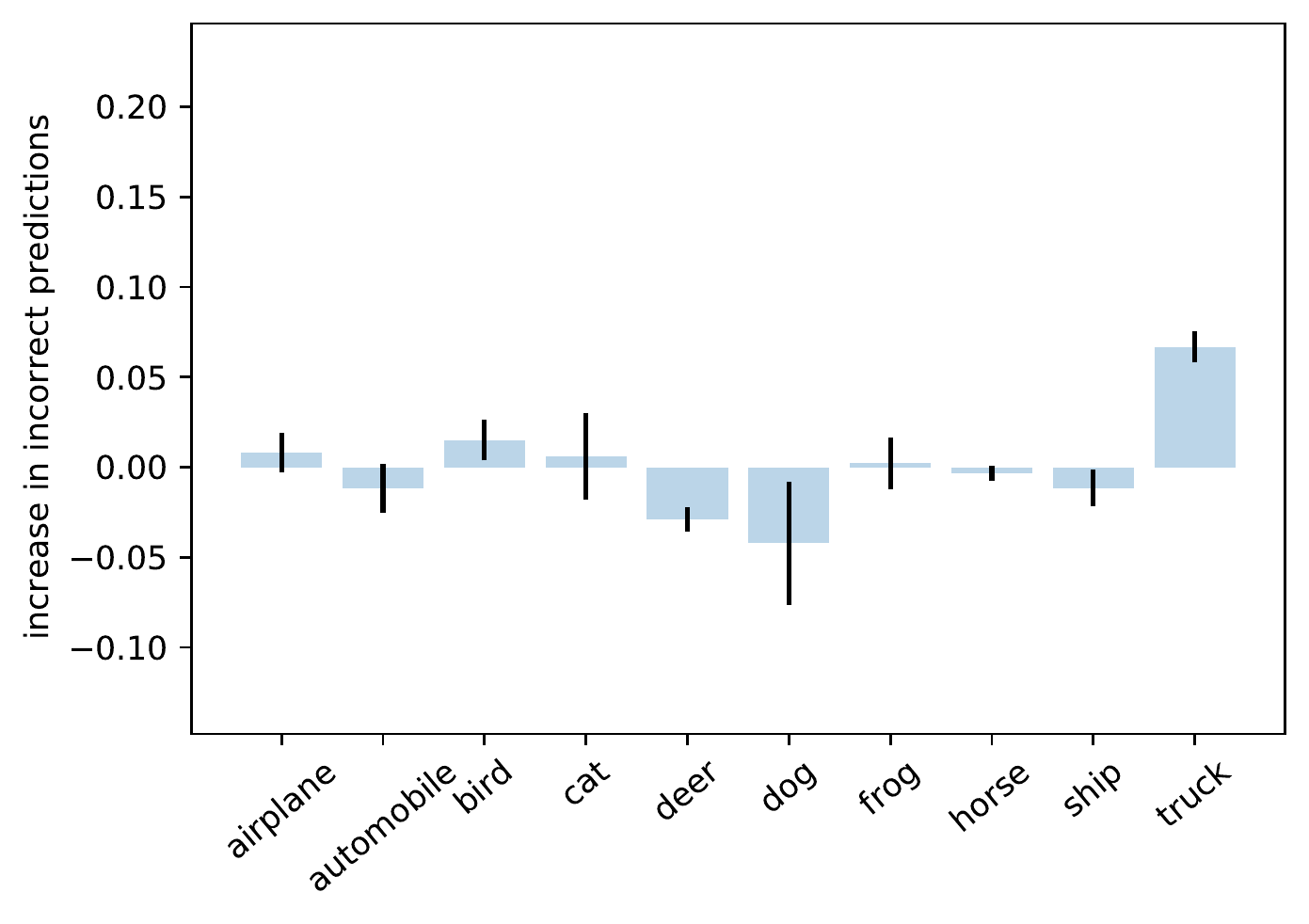}
    \end{subfigure}
    \caption{Distribution of the increase in wrong predictions when evaluating on \cutmix images for the model trained with one random mask (left) and three random masks (right).}
    \label{fig:RMbias}
\end{figure*}

When training with either one or three fixed random masks sampled from Fourier space, the \cifar{10}-trained models are still predominantly predicting \cutmix images as ``Truck'', as depicted in Figure\ref{fig:RMbias}. Again, the superimposed patches are taken from \cifar{100} images.

\end{document}